%% file: main.tex
\PassOptionsToPackage{dvipsnames}{xcolor}
\documentclass[11pt,letterpaper]{article}

\usepackage[utf8]{inputenc} %
\usepackage[T1]{fontenc}    %
\usepackage{xcolor}         %
\providecolor{niceRed}{RGB}{190,38,38}
\providecolor{niceYellow}{HTML}{f5b400}
\providecolor{blueGrotto}{HTML}{059DC0}
\providecolor{royalBlue}{HTML}{057DCD}
\providecolor{navyBlue}{HTML}{0B579C}
\providecolor{yaleBlue}{HTML}{00356b}
\providecolor{limeGreen}{HTML}{81B622}
\providecolor{nicePurple}{HTML}{9c27b0}
\providecolor{lightRoyalBlue}{HTML}{def2ff}
\providecolor{gold}{HTML}{ffa300}
\usepackage{times}
\usepackage{booktabs}       %
\usepackage{nicefrac}       %
\usepackage{microtype}      %

\usepackage[sc]{mathpazo}

\usepackage{hyperref}
\hypersetup{
  colorlinks = true,
  urlcolor = {blueGrotto},
  linkcolor = {royalBlue},
  citecolor = {navyBlue}
} 

\usepackage[margin=1in]{geometry}
\usepackage[english]{babel}

\usepackage{cmap}
\usepackage[T1]{fontenc}
\usepackage{bm}

\usepackage{amsmath}
\usepackage{amsfonts}
\usepackage{amssymb}
\usepackage{amsbsy}
\usepackage{amsthm}
\usepackage{dsfont}

\usepackage{mathtools}
\usepackage{xspace}
\usepackage{mleftright,xparse}

\usepackage{graphicx}

\usepackage{subcaption}
\usepackage{rotating}
\usepackage{float}
\usepackage{tikz}

\usepackage{algorithm}
\usepackage[noend]{algpseudocode}
\usepackage{listings}

\usepackage{enumitem}

\usepackage{multirow}
\usepackage{array}

\usepackage{accents}

\usepackage[outline]{contour}%
\usepackage{xcolor}

\usepackage[%
linewidth=2pt,
linecolor=gray,
middlelinecolor= black,
middlelinewidth=0.4pt,
roundcorner=1pt,
topline = false,
rightline = false,
bottomline = false,
rightmargin=0pt,
skipabove=0pt,
skipbelow=0pt,
leftmargin=0pt,
innerleftmargin=4pt,
innerrightmargin=0pt,
innertopmargin=0pt,
innerbottommargin=0pt,
]{mdframed}

\usepackage{chngcntr}
\usepackage{soul}
\usepackage{arydshln}

\usepackage{thm-restate}
\usepackage[capitalise,noabbrev,nameinlink,sort]{cleveref}
 
\usepackage{comment}
\usepackage[suppress]{color-edits}
\addauthor[Anay]{am}{gold}
\addauthor[Manolis]{mz}{niceRed}
\addauthor[Phuc]{phuc}{blue}
\addauthor[Van]{vv}{orange}
\usepackage[
  backref=true,
  backend=biber,
  natbib=true,
  style=alphabetic,
  sorting=alphabeticlabel,
  sortcites=true,
  minbibnames=3,
  maxbibnames=999,
  mincitenames=4,
  maxcitenames=4,
  minalphanames=4,
  maxalphanames=4,
  doi=false,
  isbn=false,
]{biblatex}

\DeclareSortingTemplate{alphabeticlabel}{
  \sort[final]{%
    \field{labelalpha}
  }
  \sort{%
    \field{year}
  }
  \sort{%
    \field{title}
  }
}

\AtEveryBibitem{%
  \clearname{editor}%
}
\AtEveryBibitem{%
  \clearlist{location}%
}

\AtBeginRefsection{\GenRefcontextData{sorting=ynt}}
\AtEveryCite{\localrefcontext[sorting=ynt]}
\input{./headers/head.tex}

\newcommand{\Ber}{\mathrm{Ber}}
\renewcommand{\wh}[1]{\smash{\widehat{#1}}}
\newcommand{\op}{\mathrm{\mathrm{op}}}
\newcommand{\obs}{\mathrm{obs}}
\newcolumntype{C}[1]{>{\centering\let\newline\\\arraybackslash\hspace{0pt}}m{#1}}

\newcommand{\highlight}[1]{\emph{\textcolor{royalBlue}{#1}}}

\newcommand{\paperTitle}{Improved Guarantees for Heterogeneous\\ Treatment-Effect Estimation via Matrix Completion}

\title{\paperTitle{}} %
\author{
        \begin{tabular}{C{7.5cm}C{7.5cm}}
        {\bf Anay Mehrotra}
            & {\bf Phuc Tran}\\
        {Stanford University}
            & {Vin University}\\[4mm]
        {\bf Van H.~Vu}
            & {\bf Manolis Zampetakis}\\
        {The University of Hong Kong}
            & {Yale University}
        \end{tabular}
}
\date{}
\begin{document}

\pagestyle{empty}
\pagenumbering{roman}
\maketitle
\thispagestyle{empty}

\begin{abstract}
A central goal of modern causal inference is estimating heterogeneous treatment effects to answer questions like ``how does an intervention affect each unit,'' rather than only on average.
We study this problem with panel-data where we observe $n$ units across $m$ times under unknown, non-uniform treatment assignments.
The data in this setting is naturally represented as a matrix of all unit--time treatment effects. 
Estimating heterogeneous treatment effects can then be expressed as obtaining a good estimation of each row's average in this matrix.
This allows us to formulate the problem as matrix completion, which can be solved under natural low-rankness assumptions.
However, existing matrix-completion guarantees are not powerful enough to get meaningful bounds for the per-row guarantee required for estimating the heterogeneous treatment effect; roughly speaking, they are only useful for estimating \textit{average} treatment effect bounds, as also illustrated in a recent line of work.
We give a simple, computationally efficient estimator that, without knowledge of the propensities and under standard low-rankness and regularity assumptions, achieves a row-wise $\ell_2$ error of $\smash{\wt{O}}(\sqrt{\nfrac{1}{n} + \nfrac{n}{(m^2)}})$.
Technically, our analysis establishes the first sharp row-wise $\ell_2$-perturbation bound for low-rank approximation, complementing \mbox{existing spectral-, Frobenius-, and entrywise perturbation theory.}

\end{abstract}

\section{Introduction}

A central question in modern causal inference is to understand for whom and
when a treatment works, not just whether it works on
average~\citep{athey2016recursive}. 
As \citet{kravitz2004evidence} put it, \emph{``the benefit or harm of [average]
treatments in clinical trials can be misleading and fail to reveal the
potentially complex mixture of substantial benefits for some, little benefit
for many, and harm for a few.''} Estimating such heterogeneous effects
requires data that observes the same units (representing, \eg{}, users,
patients, customers, or regions) across many randomized
occasions, and a range of modern experimental designs produces data of
precisely this form, including mobile-health trials, sequential A/B tests,
contextual bandit experiments, and switchback
experiments~\citep{johari2022always,li2010contextual,bojinov2023design}.
In all of these settings, we observe a panel of $n$ units  across $m$ times, where $m$ is typically larger than $n$ (\ie{}, $m\geq  n$). 
Each unit--time pair $(i,j)$ has two potential outcomes: $Y_{ij}(1)$ which arises under treatment and $Y_{ij}(0)$ which arises in the absence of treatment, \ie{}, in control.
We observe only one outcome, the one corresponding to the assignment that was actually made~\citep{imbens2015causal}. 
Stacking across units and times yields two outcome matrices $Y(1),Y(0)\in\R^{n\times m}$ in which exactly one entry from each pair $\inparen{Y_{ij}(1),Y_{ij}(0)}$ is observed.
The other entry is missing and is the counterfactual we would like to estimate.

We focus on the randomized setting, in which the treatment indicator
$D_{ij}\in\{0,1\}$ is an independent Bernoulli draw with probability
$p_{ij}\in(0,1)$. 
These probabilities can vary across units and times, reflecting different protocols, contexts, or eligibility rules. 
Further, they are, in general, unknown to the analyst: this is because, in many logged or retrospective experiments the assignment matrix is recorded but the platform's specific traffic splits, randomization rules, and ramp-up schedules are not~\citep{schnabel2016recommendations,swaminathan2015batch}.
A canonical example is the family of mobile-health trials that randomize
each participant many times over the course of an experiment, known as
\emph{micro-randomized
trials}
\citep{klasnja2015microrandomized,liao2016sample,boruvka2018assessing,qian2022microrandomized}.

As a concrete example, consider the HeartSteps study~\citep{klasnja2019heartsteps}. Here, researchers followed $n=44$ participants for six weeks, randomizing each at up to five points per day among three options (no notification, a walking suggestion, or a stand-up suggestion) and recorded the participant's step count over the following thirty minutes as the outcome; this yielded $m=210$ times per participant. The average treatment effect, aggregated across all participants and notification types, was not statistically significant. However, a finer analysis revealed striking heterogeneity: for instance, the lift from walking suggestions, which initially more than doubled step count, decayed substantially over the six weeks, while stand-up suggestions had no detectable effect throughout. Such findings translate directly into actionable interventions, for example, reducing the frequency of walking suggestions as a trial progresses, and they are visible only through estimates of \emph{heterogeneous} treatment effects.

\paragraph{Signal--Noise Model.}
A substantial line of work in causal inference is devoted to
estimating heterogeneous treatment
effects~\citep{athey2016recursive,wager2018estimation,chernozhukov2018doubleml,kunzel2019metalearners}.
To formalize the treatment effect, we adopt the standard signal-plus-noise model
$Y(a)=A(a)+E(a)$ for $a\in\{0,1\}$, where $\Ex[E(a)]=0$, so that
$A(a)\coloneqq\Ex[Y(a)]$
is the expected potential outcome under action $a$ for each unit at each
time. The expectation is taken over noise from the environment and other
unobserved factors.
The natural object for heterogeneous-effect estimation is then the unit--time
treatment-effect matrix
\[
    M\coloneqq A(1)-A(0)=\Ex[Y(1)]-\Ex[Y(0)]\,,
\]
whose row $M_i=(M_{i1},M_{i2},\ldots,M_{im})$ records how unit $i$'s
response to the treatment varies over time.
For each unit $i$, one can define several heterogeneous treatment effects by computing averages $\mathrm{Avg}_i(S)\coloneqq \smash{\sum_{j\in S}} M_{ij}/\abs{S}$ over subsets $S\subseteq [m]$.
For instance, if $S=[m]$, then $\mathrm{Avg}_i(S)$ is the treatment effect specifically for unit $i$ averaged over all times.
One can also select $S$ to be other meaningful subsets such as the first week of a trial, weekends, or
high-engagement days.

\paragraph{Matrix completion for causal inference.} 
Estimating the matrix $M$ naturally decomposes into two
estimation problems, one for $\Ex[Y(1)]$ and one for $\Ex[Y(0)]$. Fix
any $a\in\zo$: estimating $\Ex[Y(a)]$ is a matrix-completion problem,
where the goal is to recover the missing entries from those observed.
Without further structure on the signal matrices $A(0)$ and $A(1)$ this
task is impossible; however, real-world matrices have natural structural constraints that make the task
tractable. 
In particular, \citet{athey2021matrix} proposed utilizing (approximate) low-rankness of the potential output matrices.
Intuitively, low-rankness amounts to assuming that unit responses are governed by a small number of latent factors (such as baseline
activity level, receptiveness to prompts, time-of-day patterns,
habituation, or seasonality).
This assumption is well grounded in the causal panel data literature, where a
long line of work 
builds on precisely this latent-factor
structure~\citep{abadie2003economic,abadie2010synthetic,xu2017generalized,arkhangelsky2021synthetic,bai2021matrix}.
Casting the problem as matrix completion has the further benefit of
bringing to bear a broad and well-developed algorithmic
toolkit, \eg{}, \citep{candes2009exact,candes2010power,mazumder2010spectral}. %

\paragraph{Limitations of prior guarantees.}
Following~\citet{athey2021matrix}, a growing line of work has applied
matrix completion to treatment effect estimation under low-rankness and other standard regularity requirements~\citep{xu2017generalized,bai2021matrix,arkhangelsky2021synthetic,agarwal2023causal}.
The guarantees produced by these works, however, are not strong enough
for our purpose:
\begin{itemize}[leftmargin=15pt,itemsep=-1pt,topsep=-2pt]
        \item They either control the error in
    Frobenius norm,
    $\snorm{\wh{M}-M}_{\mathrm{F}}/\sqrt{nm}$, as in~\citet{athey2021matrix}.
    This is insufficient to control the above heterogeneous effects $\mathrm{Avg}_i(S)$ because, \eg{}, a single unit $i$ with constant row-normalized error
    $\norm{\wh{M}_i-M_i}_2/\sqrt{m}=\Omega(1)$ contributes only
    $O(\nfrac{1}{\sqrt{n}})$ to the normalized Frobenius error, and so leaves
    $\mathrm{Avg}_i(S)$ uncontrolled at that unit.   
    \item Or they control the worst-case entrywise error $\max_{i,j}\sabs{\wh{M}_{ij}-M_{ij}}$, as in~\citet{agarwal2023causal}. While this is strong enough to bound $\mathrm{Avg}_i(S)$, the best known bounds on this entrywise error decay only as $\nfrac{1}{\sqrt{\log m}}+\nfrac{1}{\sqrt{\log n}}$~\citep{agarwal2023causal}, which is prohibitively slow.\footnote{They incur this slow rate because their rate depends on the size of the largest all-1s sub-matrix of $D$ which, in our setting, has sides $O(\sqrt{\log{n}})$ and $O(\sqrt{\log{m}})$ with high probability.} 
    Indeed, achieving an error of, \eg{}, $0.1$ requires an astronomically large $n,m\gtrsim e^{100}$.
    
\end{itemize}
Thus, in our setting, none of the existing guarantees yields a meaningful bound.  
Let $\wh{M}$ be an estimator of $M$. 
A natural way to estimate $\mathrm{Avg}_i(S)$ for each unit $i$ and subsets $S$ of interest is to satisfy: %
\[
    (\nfrac{1}{\sqrt{m}})\cdot\snorm{M-\wh{M}}_{2,\infty}
    \coloneqq 
    (\nfrac{1}{\sqrt{m}})\cdot\max\nolimits_{1\leq i\leq n}\, \snorm{M_i-\wh{M}_i}_2
    \leq \poly\sinparen{\nfrac{1}{n},\nfrac{1}{m}}\,.
    \yesnum\label{eq:desired-guarantee}
\]
This immediately gives bounds on the estimation error of
$\mathrm{Avg}_i(S)$.
Let $\smash{\widehat{\mathrm{Avg}}}_i(S)\coloneqq\sum_{j\in S}\wh{M}_{ij}/\abs{S}$.
Now Cauchy--Schwarz inequality shows that, for every unit $i$ and subset
$S\subseteq[m]$,
\[
    \sabs{\smash{\widehat{\mathrm{Avg}}}_i(S)-\mathrm{Avg}_i(S)}
    ~\leq~
    \sqrt{\nfrac{1}{\abs{S}}}
    \cdot
    \snorm{M-\wh{M}}_{2,\infty}
    ~\leq~
    \sqrt{\nfrac{m}{\abs{S}}}\cdot \poly(\nfrac{1}{n},\nfrac{1}{m})
    \,.
\]
Hence, for any unit $i$ and every subset $S$ with $\abs{S}=\Omega(m)$, $\sabs{\smash{\widehat{\mathrm{Avg}}}_i(S)-\mathrm{Avg}_i(S)}\leq\poly(\nfrac1n,\nfrac{1}{m})$.

This leads us to the central question in the paper: 
Can such a row-wise guarantee be achieved from the observed entries, under existing structural assumptions? A bit more formally:

\vspace{2mm}
\begin{center}
\begin{mdframed}
\textbf{Question.}
    \textit{Given the observed outcomes, can one efficiently compute a matrix $\wh{M}$ that, under existing low-rankness and regularity conditions, satisfies $(\nfrac{1}{\sqrt{m}})\snorm{M-\wh{M}}_{2,\infty}\leq \poly(\nfrac1n,\nfrac1m)$?
    }
\end{mdframed}
\end{center}

\subsection{Our contributions} 
\label{sec:our-contribution}

We answer the above question affirmatively, providing an estimator that
achieves $(\nfrac{1}{\sqrt{m}})\snorm{M-\wh{M}}_{2,\infty} \lesssim\sqrt{\nfrac{1}{n}+\nfrac{1}{m}}$ with high
probability under standard low-rankness and regularity conditions.
Concretely, we work under the following regularity conditions on the signal $A$ and noise $E$. 
\begin{infassumption}[Signal and noise regularity]\label{asmp:main}
For each action $a\in\zo$, let
$\sigma_1(a)\geq \sigma_2(a)\geq\dots$ denote the singular values
of $A(a)$, and let $U(a),V(a)$ collect the left and right singular vectors
corresponding to its leading $r$ singular components. The following hold for each $a\in \zo$:
\begin{itemize}[leftmargin=13pt,itemsep=-1pt,topsep=0em]
    \item \textbf{(Approximate low rankness)}
    $A(a)$ has at most $r$ ``large'' singular values: $\sigma_{r+1}(a) \lesssim K\sqrt{m+n}.$

    \item \textbf{(Bounded signal and noise)}
    The signal and noise are bounded: $\norm{A(a)}_\infty, \norm{E(a)}_{\infty}\leq K.$
    \item \textbf{(Independent and mean-zero noise)}
    The entries of $E(a)$ are independent and mean-zero.

    \item \textbf{(Row and column incoherence)}
    The leading left and right singular spaces are ``incoherent:''
    \[
        \sqrt n \max\nolimits_{1\leq i\leq n}\snorm{e_i^\top U(a)}_2 \leq \mu
        \quadand
        \sqrt m \max\nolimits_{1\leq j\leq m}\snorm{e_j^\top V(a)}_2 \leq \mu\,.
    \]
    \item \textbf{(Signal-to-noise ratio)}
    The first singular value of $A$ is ``large:'' $\sigma_{1}(a)\gtrsim rK\sqrt{n+m}$.
\end{itemize}
\end{infassumption}
As mentioned, approximate low-rankness is standard and well motivated \cite{athey2021matrix}.
Further, boundedness and mean-zero noise are mild regularity requirements satisfied in most practical settings.
The incoherence and signal-to-noise assumptions are regularity assumptions that can also be shown to be necessary.
Row incoherence rules out the degenerate case in which the signal is concentrated on a few units; and, due to this, the information from these units' rows is not reliably available from the rest of the rows.
Column incoherence is the analogous requirement across time. 
Finally, the lower bound on the signal-to-noise ratio is mild and needed for the algorithm to reliably estimate the rank $r$ from the observed data; it is widely used in matrix perturbation work \citep{SS1}.

We need some additional notation to state our result. 
For each action
$a\in\zo$, define $p_{ij}(a)\coloneqq \Pr(D_{ij}=a),$
so that $p_{ij}(1)=p_{ij}$ and $p_{ij}(0)=1-p_{ij}$. Let
$\overline{p}_i(a)\coloneqq (\nfrac{1}{m})\,\sum_{j=1}^m p_{ij}(a)$
be the average probability of observing $Y(a)$ for unit $i$. We define
\[
    q \coloneqq \min\nolimits_{a, i}~ \overline{p}_i(a),
    \quad
    r_p \coloneqq
    \max\nolimits_{a,i,j}~
    \frac{p_{ij}(a)}{\overline{p}_i(a)},
    \quadand
    P(a)_{ij}\coloneqq 
        \frac{p_{ij}(a)}{\overline{p}_i(a)}
        - 1\,.
    \yesnum\label{def:intro_q_rp_P}
\]
Here, $q$ is the smallest average observation rate for any unit $i$ and action $a$. 
The parameter
$r_p$ measures how uneven the observation probabilities can be within a
row after normalizing by the row average. Finally, $P(a)$ measures the
remaining within-row nonuniformity of the design. 
To gain some intuition, consider the special case where all observation probabilities are constant, $p_{ij}=c$.
In this case, $q=c$, $r_p=1$, and $P(a)=0$ for each $a$.

\begin{theorem}[Main guarantee, informal]\label{thm:intro-main}
Under \cref{asmp:main}, there is a polynomial-time algorithm which, given
only the observed outcomes $Y_{\rm obs}$ and the assignment matrix $D$, and
\emph{without any knowledge of the propensities $p_{ij}$}, outputs an
estimate $\wt{M}$ satisfying, with high probability, 
\[
    \frac{1}{\sqrt{m}}\cdot\snorm{M-\wh{M}}_{2,\infty}
    ~\leq~
    \wt{O}\!\left(
        K r^{3/2}\mu
        \left[
            \sqrt{\frac{r_p}{q}\left(\frac1n+\frac{n}{m^2}\right)}
            +
            \frac{\max_{a}\norm{P(a)}_{\op}}{\sqrt{m \cdot \min\{m,n\}}}
        \right]
    \right).
\]
\end{theorem}
In the simplest case where $p_{ij}=c$ (for each $i,j$) and $K,r,\mu=O(1)$, the main theorem yields

\[
    \frac{1}{\sqrt{m}}\cdot\snorm{M-\wh{M}}_{2,\infty}
    ~\leq~
    \wt{O}\!\left(
        \sqrt{\frac1n+\frac{n}{m^2}}
    \right)\,,
\]
which achieves the desired $\poly(\nfrac1n,\nfrac1m)$ rate in the standard regime where $m\gtrsim n$. 
In contrast, in the same setting the prior works either do not provide any non-trivial bound \citep{athey2021matrix} or only achieve a $\poly(\nfrac{1}{{\log n}},\nfrac{1}{{\log m}})$ rate \cite{agarwal2023causal}. %

Another useful special case is the \emph{row-homogeneous design}, where
$p_{ij}=p_i$ for all $j$. This models experiments in which different units
may be randomized at different rates, but each unit's randomization rate is
stable over time. 
Here, we obtain the following corollary

\begin{corollary}[Row-homogeneous propensities, informal]\label{cor:intro-row-hom}
Suppose \cref{asmp:main} holds, $p_{ij}=p_i$ with
$p_i\in[\Omega(1),1-\Omega(1)]$, and $K,r,\mu=O(1)$. Then with high
probability,
$
 (\nfrac{1}{\sqrt{m}})\cdot\snorm{M-\wh{M}}_{2,\infty}
    ~\leq~
    \wt{O}(
        \sqrt{\nfrac1n+\nfrac{n}{m^2}}
    ).$
\end{corollary}

\paragraph{Technical Novelty (also see \cref{sec: tech overview}).}
One important advantage of the estimator that we propose is that it is very simple and easy to implement: it is a simple row-scaled truncated-SVD algorithm
(\cref{alg:row-scaled-svd}). 
The analysis of this simple algorithm in our setting though, requires several new tools in the spirit of modern results from matrix perturbation theory.
Existing work on perturbation bounds of low-rank approximations
has primarily focused either on bounds for different norms \eg{},
\cite{tran2025newlowrank, TranVishnoiVu2025,EY1,ltranvufast, chatterjee2015matrix,MangoubiVJACM, Dwok2, AbVuinfinity}, which are not applicable in our setting, or bounds that do not gracefully improve for low-rank matrices
\citep{cape2019two,abbe2020entrywise}, and hence, are not useful for analyzing our algorithm.
\highlight{Our main technical contribution is a perturbation theory for
truncated SVD in the $\norm{\cdot}_{2,\infty}$ norm, which directly controls
the reconstructed error of our algorithm.}
To obtain the required sharp bound, we need to carefully adapt the contour expansion method, introduced in \cite{DKTranVu, ltranvufast, TranVishnoiVu2025} and applied in the more challenging norm $\norm{\cdot}_{2,\infty}$. In particular, the contour bootstraping argument used in \cite{DKTranVu, ltranvufast, TranVishnoiVu2025} does not apply in our case and we need a new idea to control the error that arises from the contour expansion method. We refer the reader to \cref{sec: tech overview} for more details. %

A second difficulty comes from the fact that the probabilities $p_{ij}$ are
unknown and nonuniform.  If the entrywise probabilities $p_{ij}(a)$ were
known, one could form an unbiased inverse-propensity-weighted matrix by
scaling each observed entry by $1/p_{ij}(a)$.  Our algorithm does not have
this information and instead scales row $i$ by its empirical observation
frequency.  The population analogue of this row scaling is unbiased when the
design is row-homogeneous (\ie{}, $p_{ij}=p_{ik}$ for each $j,k$), but under nonuniform propensities its expectation
satisfies
$\sfrac{p_{ij}(a)A_{ij}(a)}{\overline p_i(a)}
    =
    A_{ij}(a)+P_{ij}(a)A_{ij}(a).$
Thus the perturbation is not purely mean-zero noise: it also contains a
deterministic bias induced by nonuniformity within one row.  Our analysis
incorporates this bias directly into the perturbation argument, yielding a
bound whose additional design-dependent term is governed by
$\norm{P(a)}_{\op} / \sqrt{m \cdot \min\sinparen{m,n}}$, which decays as $m,n\to\infty$ and vanishes when $P(a)=0$.

\subsection{Additional Related Works} 
\label{sec:related-work}

Our work is broadly related to two lines of work: (1) work utilizing low-rank
approximation results in causal inference and (2) work on matrix perturbation
bounds for low-rank approximations.
We have already mentioned key works from both lines in the introduction.
Here we give a more detailed comparison with the first line.
A more technical comparison with the second line appears in Technical Overview (\cref{sec: tech overview}).
A growing literature uses low-rank structure to estimate missing counterfactual
outcomes in panel data.
As we have already mentioned, closest to our work are \citet{athey2021matrix,agarwal2023causal}, and their guarantees are insufficient for the heterogeneous effects we study.
Several other works also exploit low-rankness in panel outcomes but their goal is to estimate (different types of) \textit{average} treatment effects~\citep{amjad2018robust,amjad2019mrsc,arkhangelsky2021synthetic,bai2021matrix,fernandezval2021lowrank}.
In contrast, we estimate heterogeneous treatment effects.
A different line of work imposes low-rankness on the covariate matrix (rather than on
the potential-outcome matrices) and use it to impute or de-noise covariates~\citep{agarwal2021robustpcr,agarwal2025modelpcr}, while they also use low-rankness, their estimation targets and setting is quite different from our work.
Finally, \citet{agarwal2025syntheticinterventions} extend the low-rank
framework to settings with more than two treatment conditions and give
row-wise guarantees for estimating $\mathrm{Avg}_i([m])$.
Our row-wise control of $M$ is stronger: it yields bounds on
$\mathrm{Avg}_i(S)$ for every sufficiently large $S\subseteq[m]$, not just
$S=[m]$.

\section{Preliminaries}

\label{sec:preliminaries}
    In this section, we introduce basic notation and preliminaries.
    
\paragraph{Notation.}
For constants $N$ and $1\leq k\leq N$, let $e_{N,k}$ be the $k$th standard basis vector in $\R^N$; when the dimension is clear, we write $e_k$.
For a vector $v$ and $p\geq 1$, define $\norm{v}_p$ as $(\sum_{j}\abs{v_j}^p)^{1/p}$ and $\norm{v}_\infty$ as $\max_j\abs{v_j}$.
For a matrix $B\in\R^{n\times m}$, we write $B_i$ for its $i$th row, $B_{\cdot j}$ for its $j$th column, $\norm{B}_\infty$ for $\max_{i,j}\abs{B_{ij}}$, and $\norm{B}_{\op}$ and $\norm{B}_{\mathrm{F}}$ for its operator and Frobenius norms respectively.
The central object for our row-wise guarantees is the row-wise $\ell_2$-norm $\norm{B}_{2,\infty}\coloneqq\max\nolimits_{1\leq i\leq n}\norm{B_i}_2$.

\paragraph{Potential Outcomes and the Heterogeneous Treatment Effect.}
We observe $n$ units (\eg{}, users or patients) across $m$ times (\eg{}, decision points or days).  For each unit--time pair $(i,j)$ and action $a\in\zo$, $Y_{ij}(a)$ denotes the potential outcome of unit $i$ at time $j$ under action $a$.  Following the standard potential-outcomes framework~\citep{imbens2015causal,hernan2023causal}, we adopt the signal-plus-noise model $Y(a)=A(a)+E(a)$ with $\E[E(a)]=0$, so that $A(a)\coloneqq\E[Y(a)]$ is the mean potential-outcome matrix.  Our goal is to estimate the heterogeneous treatment-effect matrix $M\coloneqq A(1)-A(0)$, whose entry $M_{ij}$ is the mean effect of treating unit $i$ at time $j$.  Unless otherwise stated, expectations are taken over the outcome noise $E$, not the treatment assignments.

\paragraph{Assignment and Observed Data.}
We make standard assumptions on the assignment mechanism \cite{hernan2023causal,imbens2015causal}.
For each $(i,j)$, the treatment indicator $D_{ij}\in\{0,1\}$ is an independent Bernoulli draw, $D_{ij}\sim\Ber(p_{ij}),$ where the probabilities $p_{ij}$ may vary across units and times and are not assumed to be known. 
The assignments $D=(D_{ij})$ are independent of the potential outcomes $\{Y(0),Y(1)\}$, and the observed outcome at $(i,j)$ is $Y^{\obs}_{ij}\coloneqq D_{ij}Y_{ij}(1)+(1-D_{ij})Y_{ij}(0).$
To treat the two actions symmetrically, we introduce, %
\[
    D_{ij}(1)\coloneqq D_{ij}\,,\quad D_{ij}(0)\coloneqq 1-D_{ij}\,,
    \quad
    p_{ij}(1)\coloneqq p_{ij}\,,\quad p_{ij}(0)\coloneqq 1-p_{ij}\,,
\]
so that $D_{ij}(a)=1$ when $Y_{ij}(a)$ is observed and $\E[D_{ij}(a)]=p_{ij}(a)$. 
The corresponding partially observed matrix $\wt{Y}(a)$ is defined entrywise by
\[
    \wt{Y}_{ij}(a) \coloneqq Y_{ij}(a)\;\quadtext{if} D_{ij}(a)=1\qquadand \wt{Y}_{ij}(a) \coloneqq \star\;\quad\text{otherwise}\,, 
\]
with $\star$ denoting a missing entry. The analyst equivalently sees the pair $(D,Y^{\obs})$, or the pair of partially observed matrices $(\wt{Y}(0),\wt{Y}(1))$ together with their observation masks.

\paragraph{Row Propensities and Non-uniformity.}
The rates we obtain depend on the parameters $q$, $r_p$, and $P(a)$ defined in \cref{def:intro_q_rp_P}.  
The parameter $q$ controls how often the rarest action is observed at the rarest unit.  Requiring $q\geq\Omega(1)$ is substantially milder than the standard overlap condition~\citep{imbens2015causal,hernan2023causal}, which requires every $p_{ij}$ to be bounded away from $0$ and $1$; our condition instead constrains only the row averages, leaving individual $p_{ij}$ free to be arbitrarily close to $0$ or $1$ at many time points within a unit.
The parameters $r_p$ and $P(a)$ quantify within-row nonuniformity.  By construction, each row of $P(a)$ averages to zero, and $P(0)=P(1)=0$ exactly when the design is row-homogeneous, \ie{}, $p_{ij}=p_i$ for all $i,j$.

\paragraph{SVD, Low-Rankness, and Noise.}
For a rank-$r$ matrix $A\in\R^{n\times m}$, write $A=U\Sigma V^\top=\sum_{\ell=1}^r \sigma_\ell u_\ell v_\ell^\top$, where $U=[u_1,\ldots,u_r]\in\R^{n\times r}$ and $V=[v_1,\ldots,v_r]\in\R^{m\times r}$ have orthonormal columns and the singular values $\sigma_1\geq\sigma_2\geq\cdots\geq\sigma_r$ are arranged in non-increasing order.  We write $A_s$ for the best rank-$s$ approximation of $A$, and \[\delta_s(A)\coloneqq\sigma_s-\sigma_{s+1}\] for its $s$th singular value gap; when $A$ is clear from context we abbreviate $\delta_s\coloneqq\delta_s(A)$.
The row- and column-incoherence parameters of $A$ are
\[
    \mu_R(A)\coloneqq\sqrt n\max\nolimits_{1\leq i\leq n}\snorm{e_{n,i}^\top U}_2
    \qquadand
    \mu_C(A)\coloneqq\sqrt m\max\nolimits_{1\leq j\leq m}\snorm{e_{m,j}^\top V}_2\,,
    \yesnum\label{def:incoherence}
\]
and we set $\mu(A)\coloneqq\max\{\mu_R(A),\mu_C(A)\}$. 
These quantities are small when the singular spaces spread across units and time points, and large when either side concentrates on few coordinates.
When both potential-outcome signals are under consideration we write $\mu\coloneqq\max_{a\in\{0,1\}}\mu(A(a))$.

\begin{definition}[\((K,\sigma)\)-Bounded Random Matrix]
\label{def:bounded-noise}
A random matrix $E_R\in\R^{n\times m}$ is called \((K,\sigma)\)-bounded if its entries satisfy $\E E_{R,ij}=0,$ $\E[|E_{R,ij}|^2]\leq \sigma^2,$ and $\E[|E_{R,ij}|^\ell]\leq K^{\ell-2}\sigma^2$ 
for every $\ell\geq2$ and all $i\in[n]$, $j\in[m]$.

\end{definition} 
This condition fixes the variance scale and supplies the higher-moment control used in Bernstein-type matrix concentration \citep{tropp2012user,tropp2015introduction}.

\section{Our Main Results} 

\label{sec:main-result}
\label{sec: main result}
In this section, we present our main results.
We begin with the formal version of \cref{asmp:main}, followed by our estimator, and then its guarantees.
Recall the parameters $q,r_p$, and $P(a)$ from \cref{def:intro_q_rp_P}. 
For each $a\in\zo$, write 
\[
    T(a)
    \coloneqq
    \sqrt{
        \frac{m+r_p n}{q}\,
        \log(m+n)
    }
    +
    \norm{P(a)}_{\op}\,. 
\]
We impose the following standard regularity conditions
\begin{assumption}[Regularity conditions]
\label{asmp:main-formal}
    Fix constants $K_A,K_E,\mu>0$ and $r\in \N$.
    Set $K\coloneqq K_A+K_E$.  
    For each $a\in\zo$, the following conditions hold:
\begin{enumerate}[leftmargin=15pt,itemsep=-1pt,topsep=0em]
    \item \textbf{(Low-Rankness)} 
    \mbox{$\norm{A(a)-A_r(a)}_{\op} \lesssim K\sqrt{m+n},$ where $A_r(a)$ is $A(a)$'s best rank-$r$ approximation}

    \item \textbf{(Bounded Signal and Noise)} 
    $\norm{A(a)}_\infty\leq K_A$
    and 
    $E$ is $(K_E,K_E)$-bounded (\cref{def:bounded-noise}).
    \item \textbf{(Independent and mean-zero)} \mbox{$E(a)$'s entries are independent of each other and $D$, and mean-zero}

    \item \textbf{(Row and column incoherence)} 
    $\mu_C(A(a)), \mu_R(A(a))\leq \mu$, where $\mu(\cdot)$ is defined in \eqref{def:incoherence}.
 
    \item \textbf{(Signal-to-noise ratio)} The leading singular value satisfies $\sigma_1(a) \gtrsim K r T(a) .$
\end{enumerate}
\end{assumption}
We refer the reader to \cref{sec:our-contribution,sec:additional-remarks} for discussion of this assumption.

Next, we present our estimator (see \cref{alg:row-scaled-svd}).
It computes $\wh A(0)$ and $\wh A(1)$ separately and returns $\wh M\coloneqq \wh A(1)-\wh A(0)$.  To estimate $A(a)$, we fill unobserved entries with zeros, rescale each row by its empirical observation frequency, and keep the largest singular block separated by a spectral gap.

\begin{algorithm}[htb]
\caption{Row-scaled spectral estimator}
\label{alg:row-scaled-svd}
\begin{algorithmic}[1]
\State \textbf{Input:} Observed outcomes $Y^{\obs}$, matrix $D$, $r$ from \cref{asmp:main-formal}, and thresholds $\tau_0,\tau_1$.

\For{\textbf{each} $a\in\zo$}
    \State Initialize the estimates
    $\wh{p}_i(a)\leftarrow
    \max\{m^{-1}\sum_{j=1}^m D_{ij}(a),\,m^{-1}\}$
    for all $i\in[n]$.
    
    \State For each $(i,j)\in [n]\times [m]$, set $Z_{ij}(a)\leftarrow D_{ij}(a)Y^{\obs}_{ij}$ and
    $\wt{Y}^{\mathrm{ub}}_{ij}(a)\leftarrow
    Z_{ij}(a)/\wh{p}_i(a)$.

    \State Let
    $\wt\sigma^{\mathrm{ub}}_1(a)\geq
    \wt\sigma^{\mathrm{ub}}_2(a)\geq\dots$
    be the singular values of $\wt{Y}^{\mathrm{ub}}(a)$. %

    \State Let
    $\mathcal S_a\leftarrow
    \{1\leq s\leq r:
    \wt\sigma^{\mathrm{ub}}_s(a)-
    \wt\sigma^{\mathrm{ub}}_{s+1}(a)\geq \tau_a\}$.
    
    \State Set $\wh{s}(a)\leftarrow\max\mathcal S_a$ if $\mathcal S_a\neq\emptyset$,
    and $\wh{s}(a)\leftarrow0$ otherwise.

    \State Set
    $\wh{A}(a)\leftarrow[\wt{Y}^{\mathrm{ub}}(a)]_{\wh{s}(a)}$, where $[B]_s$ denotes the best rank-$s$ approximation of $B$
\EndFor

\State \textbf{Return} $\wh{M}\leftarrow \wh{A}(1)-\wh{A}(0)$.
\end{algorithmic}
\end{algorithm}

\paragraph{Running time of \cref{alg:row-scaled-svd}.} 
The dominant cost is computing the top $r+1$ singular values and vectors of $\wt Y^{\mathrm{ub}}(a)$ for each $a\in\zo$, which suffice to determine $\wh s(a)$ and $[\wt Y^{\mathrm{ub}}(a)]_{\wh s(a)}$.  This takes $\wt{O}(nmr)$ time via Lanczos's method or randomized SVD \citep{golub2013matrix}.  All remaining steps run in $O(nm+nr)$ time, giving a total running time of $\wt{O}(nmr)$.

\paragraph{Main Result.}
Next, we state our main result, which bounds the error in estimation of $A(0)$ and $A(1)$; combining the two bounds immediately implies a bound on the estimation error for $M$.
\begin{theorem}[Upper bound on Error]
\label{thm:po-row-reconstruction}
Suppose \cref{asmp:main-formal} holds.  Run
\cref{alg:row-scaled-svd} with thresholds $\tau_a=96KT(a)$ for
$a\in\zo$.  
Then, with probability at least $1-O(\nfrac{1}{(m+n)})$, the output matrices $\wh{A}(0)$ and $\wh{A}(1)$ satisfy, for each $a\in \zo$,
\
\[
    \norm{\wh A(a)-A(a)}_{2,\infty}
    ~~\lesssim~~
    K r^{3/2}\mu ~~ 
    (\sqrt{ m + n})\log^4(m+n)
    ~~
    \left[
        \sqrt{\frac{r_p}{mq} + \frac{r_p}{nq}}
        +
        \frac{\norm{P(a)}_{\op}}{\sqrt{mn}}
    \right]\,. %
\]
\end{theorem}
Now, applying \cref{thm:po-row-reconstruction} to both $a=0,1$ and using the triangle inequality yields the following row-wise guarantee for $\wh M$.

\begin{corollary}[Row-wise recovery of the treatment-effect matrix]
\label{cor:main-rowwise-treatment}
Under the assumptions of \cref{thm:po-row-reconstruction}, with probability
at least $1-O(\nfrac{1}{(m+n)})$,%
\[
    \max_{1\leq i\leq n}
    \frac{\norm{\wh M_i-M_i}_2}{\sqrt m}
    ~~\lesssim~~
    K r^{3/2}\mu~~ \log^4(m+n)~~
    \left[
        \sqrt{
            \frac{r_p}{q}
            \left(
                \frac1n+\frac{n}{m^2}
            \right)
        }
        +
        \frac{
            \max_{a\in\zo}\norm{P(a)}_{\op}
        }{
            \sqrt{m\min\{m,n\}}
        }
    \right]. 
\]
\end{corollary}
Thus, when $m\geq n$, $K,r,\mu,r_p$ and $q^{-1}$ are constants, and the design is row-homogeneous (so $P(0)=P(1)=0$), the error scales as
$\wt O(n^{-1/2})$.
Next, to build some intuition, we highlight the bounds we obtain under two regimes that are common in panel-data applications.

\paragraph{Special Case I (Row-homogeneous design):} 
Here, $p_{ij}$ may differ across units $i$ but, for each $i$, is invariant in $j$:
\[
    p_{ij}=p_i 
    \qquad \text{for every } i\in[n] \text{ and } j\in[m].
    \yesnum\label{def:row-homogeneous}
\]
This setting captures stratified and covariate-adaptive randomization protocols, and also includes the Bernoulli design $p_{ij}=c$ as a sub-case.  
It is useful because under any row-homogeneous design, $P(0)=P(1)=0$ and $r_p=1$, leading to the following bound.

\begin{corollary}[Row-homogeneous design]
\label{cor:row-homogeneous}
Consider the row-homogeneous design above (\cref{def:row-homogeneous}).
Suppose the assumptions of \cref{thm:po-row-reconstruction} hold, $K,r,\mu=O(1)$, $q=\Omega(1)$, and $m\geq n$.
Then, with probability $1-O(\nfrac{1}{(m+n)})$ it holds that $\inparen{\sfrac{1}{\sqrt{m}}} \cdot \norm{\wh M-M}_{2,\infty}
    \leq 
    \wt{O}\!\inparen{\nfrac{1}{\sqrt{n}}}$.
\end{corollary}
In the balanced regime $m\asymp n$, this matches the best known guarantees for matrix completion under the Bernoulli design, despite the fact that our algorithm does not have access to the propensities $p_i$.

\paragraph{Special Case II (Spectrally small within-row non-uniformity):} 
Our next regime is more general: it allows $p_{ij}$ to vary in time within rows but requires the variation to be \emph{spectrally} small.  Concretely, 
\[
    \text{for some $\nu\geq 0$\,,}\quad
    \max\nolimits_{a\in\zo}\norm{P(a)}_{\op}
    \leq 
    \nu ~ \wt{O}(\sqrt m+\sqrt n)\,.
    \yesnum\label{def:small-nonuniform} %
\]
For example, this holds whenever $P(a)$'s entries are independent and sub-Gaussian, as in contextual-bandit-style adaptive designs in which propensities depend sufficiently mildly on observed covariates; standard random-matrix bounds then give the spectral condition with high probability.

\begin{corollary}[Spectrally small within-row nonuniformity]
\label{cor:spectral-nonuniformity}
    Consider the setting above (\cref{def:small-nonuniform}).
    Suppose the assumptions of \cref{thm:po-row-reconstruction} hold, $K,r,\mu=O(1)$, $q=\Omega(1)$, and $m\geq n$.  
    Then, with probability at least $1-O(\nfrac{1}{(m+n)})$, it holds that $\inparen{\nfrac{1}{\sqrt{m}}} \cdot \norm{\wh M-M}_{2,\infty} \leq \nu\, \wt O(n^{-1/2})$.
    
\end{corollary}
Thus, \cref{cor:spectral-nonuniformity} preserves the $\wt O(n^{-1/2})$ row-normalized rate of the row-homogeneous case, with $\nu$ inflating only the constant. 
Crucially, the algorithm itself does not change between the two regimes; the analyst does not need to verify the spectral bound or estimate $\nu$.

\section{Technical Overview} \label{sec: tech overview}
 
In this section, we sketch the key ideas behind the proof of \cref{thm:po-row-reconstruction}. %
In this section, we focus on $a=1$ and, hence, omit it from the notation, writing, \eg{}, $A$ and $P$ for $A(1)$ and $P(1)$ respectively.
Now, our goal is to prove that with probability at least $1-O(\nfrac{1}{(m+n)}),$
\begin{equation} \label{bound: toprove}
\|\hat{A}-A\|_{2,\infty} ~~\lesssim~~ K r^{3/2} \mu  ~\log^4(m+n) (\sqrt{m}+\sqrt{n}) \inparen{ \sqrt{\frac{r_p}{mq} + \frac{r_p}{nq} }   + \frac{\|P\|_{\op}}{\sqrt{mn}} }\,.
\end{equation}
Where the constants $K$, $r$, and $\mu$ are from \cref{asmp:main-formal} and the parameters $(r_q,q,P)$ are as defined in \cref{def:intro_q_rp_P}.
Before presenting the proof, we need to set up some notation.

\paragraph{Notation and basic observations.} 
We use $A^{ub}$ to denote $\Ex \tilde{Y}^{ub}$, where $\tilde{Y}^{ub}$ is from \cref{alg:row-scaled-svd}.
A direct computation shows $A^{ub} \coloneqq  \left(\sfrac{p_{ij} a_{ij}}{p_i} \right)_{\substack{ij}}.$ 
For any $X \in \mathbb{R}^{n \times m}$, we denote the corresponding ``observed'' matrix by $\smash{\tilde{X}}=\left(\tilde{x}_{ij} \right)_{ij}$ and the corresponding ``scaled'' matrix by $\smash{\tilde{X}}^{ub} \coloneqq  \left(\tilde{x}_{ij}/p_i \right)_{ij}.$
With this notation, since $Y = A+E$, we can write 
$\tilde{Y}^{ub} = \tilde{A}^{ub}+ \tilde{E}^{ub}. $  
Finally, we define
$$E_0 \coloneqq  A^{ub} -A = \inparen{\inparen{\frac{p_{ij}}{p_i}-1} a_{ij}}_{ij} \qquadand E_R \coloneqq  (\tilde{Y}^{ub} - A^{ub}) + \tilde{E}^{ub}\,.$$
Next, we observe that \cref{asmp:main-formal}(5) and the choice that $\textstyle \textstyle \tilde{\sigma}^{\mathrm{ub}}_s
-
\tilde{\sigma}^{\mathrm{ub}}_{s+1}
\gtrsim
K T(a) $ are equivalent to 
$\sigma_1 > 10r(\|E_R\|_{\op}+ \|E_0\|_{\op}),$ and $\tilde{\sigma}^{\mathrm{ub}}_s
-
\tilde{\sigma}^{\mathrm{ub}}_{s+1}
\geq 8(\|E_R\|_{\op}+ \|E_0\|_{\op})$ respectively.
(Where we selected the constants to simplify exposition.)

\paragraph{Our approach.}
At a high-level, to prove \eqref{bound: toprove}, we split $\|\wh{A} -A\|_{2, \infty}$ into two parts:
\begin{enumerate}%
    \item \textit{Term 1 (Tail-error of low-rank approximation):}\quad $\|A_s -A\|_{2, \infty}$; and 
    \item \textit{Term 2 (Perturbation of low-rank approximations):}\quad $\|\wh{A} - A_s\|_{2, \infty}$
\end{enumerate}
One can bound the first term, $\|A_s -A\|_{2,\infty}$, by combining the singular decomposition of $A$ and the definition of the threshold position $s$. 
The key difficulty is bounding the second term, which measures the perturbation of low-rank approximations in $\|\cdot\|_{2,\infty}$.
Concretely, the triangle inequality implies
$$\|\hat{A} - A\|_{2,\infty}~~=~~\|(\hat{A} -A_s)+ (A_s -A)\|_{2,\infty} ~~\leq~~ \|\hat{A} -A_s\|_{2,\infty} + \|A_s -A\|_{2,\infty}.$$

\paragraph{Step 1 (Bounding Term 1).} 
Since $A=\sum_{i=1}^r \sigma_i u_i v_i^\top$,  $A-A_s= \sum_{i=s+1}^r \sigma_i u_i v_i^\top$. 
Therefore, 
\begin{align*}
    \|A_s -A\|_{2, \infty} 
    &~~=~~\max_{1 \leq k \leq n} \|e_{n,k}^\top (A_s-A)\|_2\\
    &~~=~~ \max_{1 \leq k \leq n} \| \ \sum\nolimits_{r\geq i>s} \sigma_i u_{ik} v_i^\top\|_2 \\
    &  ~~=~~ \max_{1 \leq k \leq n} \sqrt{\ \sum\nolimits_{i=s+1}^r \sigma_i^2 u_{ik}^2 }\\
    &~~\leq~~ \max_{s < i \leq r} \|u_i\|_{\infty}  \sqrt{\sum\nolimits_{i=s+1}^r \sigma_i^2}\\
    &~~\leq~~  \max_{s < i \leq r} \|u_i\|_{\infty}   \sqrt{r} \sigma_{s+1}\,. 
\end{align*}
We claim that $\sigma_{s+1} \le 10r(\|E_R\|_{\op}+\|E_0\|_{\op})$. 
Suppose, for contradiction, that
\(
\sigma_{s+1} > 10r(\|E_R\|_{\op}+\|E_0\|_{\op}).
\)
Since $\rank(A)\le r$ (\ie{}, $\sigma_{r+1}=0$), it follows that $s+1\le r$. 
Arguing as in Remark~\ref{rem:gap-exists} (with $\sigma_{s+1}$ in place of $\sigma_1$), there exists some $j$ with $s+1\le j\le r$ such that
\(
\tilde{\sigma}^{ub}_j-\tilde{\sigma}^{ub}_{j+1}
>
8(\|E_R\|_{\op}+\|E_0\|_{\op}).
\)
This contradicts the definition of $s$ as the largest index satisfying
\(
\tilde{\sigma}^{ub}_s-\tilde{\sigma}^{ub}_{s+1}
>
8(\|E_R\|_{\op}+\|E_0\|_{\op}).
\)
Hence, $\sigma_{s+1} \le 10r(\|E_R\|_{\op}+\|E_0\|_{\op})$.
Therefore, 
$$\|A_s -A\|_{2, \infty} \leq 10 r^{3/2} (\|E_R\|_{\op}+\|E_0\|_{\op}) \cdot \max\nolimits_{s < i \leq r} \|u_i\|_{\infty}.$$

To bound $\|E_R\|$, we use the following lemma. Its proof will be presented later in \cref{subsec: proof ER}.
\begin{lemma} \label{lem: ER bound}
   With high probability, 
   $\|E_R\|_{\mathrm{op}} \leq 12K \sqrt{\log(m+n)} \cdot \sqrt{\nfrac{r_p(m+n)}{q}}.$
\end{lemma}
We also have $\|E_0\|_{\mathrm{op}} \leq K_A \|P\|_{\mathrm{op}}.$ By the definition of $\mu$, $\max_{s < i \leq r} \|u_i\|_{\infty} \leq  \mu(\nfrac{1}{\sqrt{m}}+\nfrac{1}{\sqrt{n}})$.  Combining all estimates, we obtain 
\begin{equation} \label{As-A}
 \|A_s -A\|_{2, \infty} \leq 12\sqrt{\log (m+n)}  K r^{3/2} \mu \left(\sqrt{m}+\sqrt{n}\right) \cdot \left(\sqrt{\frac{r_p}{mq}+\frac{r_p}{nq}} + \frac{\|P\|_{\op}}{\sqrt{mn}}  \right)    .    
\end{equation}

\paragraph{Step 2 (Bounding Term 2).}
Our estimate is based on the following theorem, which is a key part of our technical contribution. 
We compare its proof techniques with prior work  in \cref{sec:discussionThmmainRec}.
The discussion of the perturbation of low-rank approximations in $\|\cdot\|_{2,\infty}$ and the detailed proof of \cref{theo: mainRec} appear in \cref{sec: lowrankresult}. 
\begin{theorem} \label{theo: mainRec}
    Let $\tilde{A}=A+E_R+E_0$, where $E_R$ is $(K,\sigma)$-bounded. There is a universal constant $C >0$ satisfying: If $\delta_s \geq 6(\|E_R\|_{\op}+\|E_0\|_{\op})$, then with probability $1-O(\nfrac{1}{(m+n)})$,
    \begin{align*}
        \frac{\|\tilde{A}_s-A_s\|_{2,\infty}}{C\sqrt{r}} \leq     \inparen{\log^2(m+n)+ \frac{K \log^4(m+n)}{\sqrt{m+n}}} \inparen{\frac{\mu}{\sqrt{m}}+ \frac{\mu}{\sqrt{n}} }  {  \frac{\sigma_s\sqrt{m+n} \inparen{\sigma+\|E_0\|_{\op} }}{\delta_s}} \,.     
    \end{align*}
\end{theorem}
Since $\tilde{Y}^{ub}=A+E_R+E_0$, we have the following observations. 
\begin{itemize}[itemsep=-1pt,topsep=0em,leftmargin=15pt]
    \item Since $\tilde{\sigma}_s^{ub} - \tilde{\sigma}_{s+1}^{ub} \geq 8(\|E_R\|_{\op}+\|E_0\|_{\op})$, by Weyl's inequality (\cref{theo: weyl}), we have 
    $$ \sigma_{s}-\sigma_{s+1}~~\geq~~ \tilde{\sigma}_s^{ub} - \tilde{\sigma}_{s+1}^{ub} - 2(\|E_R\|_{\op}+\|E_0\|_{\op}) ~~\geq~~ 6(\|E_R\|_{\op}+\|E_0\|_{\op}) .$$
    \item By the definition of $E_R=( E_{Rij})_{ij}$, we have 
    \begin{equation*}
     E_{Rij}= \begin{cases}
        &\textstyle 
        \nfrac{(1-p_{ij})a_{ij}}{p_i}  + \nfrac{\varepsilon_{ij}}{p_i}   \,\,\text{with probability }\,p_{ij}\,, \\
        &\nfrac{-p_{ij}a_{ij} }{p_i}  \,\,\text{with probability }\,1- p_{ij}.
     \end{cases}   
    \end{equation*}
\end{itemize}
For each $\ell\geq 2$, a direct moment calculation gives
\[
    \Ex\!\left[|E_{Rij}|^\ell\right]
    ~~\leq~~
    \frac{p_{ij} (K_A+K_E)^\ell}{p_i^\ell}
    ~~=~~
    \frac{p_{ij}K^\ell}{p_i^\ell}\,.
\]
    These observations allow us to apply \cref{theo: mainRec} on the pair $(\tilde{Y}^{ub}, A)$ with the noises $E_R, E_0$.  Indeed, the parameters $K$ and $\sigma$ in \cref{theo: mainRec} are replaced respectively by $\sfrac{K}{q}$ and $\sqrt{\sfrac{r_p K}{q}}$ in our setting.
    Thus, with probability at least $1 -O( (m+n)^{-1}),$ there is a constant $C>0$ such that
    \begin{equation*}
        \begin{split}
      \frac{\|\hat{A} - A_s\|_{2,\infty}}{ C\sqrt{r}}  ~~=~~ \frac{\|\tilde{Y}^{ub}_s - A_s\|_{2,\infty}}{ C\sqrt{r}}
       & ~~\lesssim~~
       \log^4(m+n) K \mu (\sqrt{m}+\sqrt{n})\frac{\sigma_s}{\delta_s} \cdot \insquare{\sqrt{\frac{r_p}{mq} + \frac{r_p}{nq} }+ \frac{\|P\|_{\op}}{\sqrt{mn}} }   
        \end{split}
    \end{equation*}
Since $\textstyle \sfrac{\sigma_s}{\delta_s} =1 + (\sfrac{\sigma_{s+1}}{\delta_s}) \leq 1 + \frac{r(\|E_R\|_{\op}+\|E_0\|_{\op})}{8(\|E_R\|_{\op}+\|E_0\|_{\op}\|)} \leq 2r,$ we further have
\begin{equation}\label{hatA-As}
   \|\hat{A} - A_s\|_{2,\infty}  ~~\leq~~ O \inparen{\log^4(m+n) K r^{3/2} \mu (\sqrt{m}+\sqrt{n}) \cdot \insquare{\sqrt{\frac{r_p}{mq} + \frac{r_p}{nq} }+ \frac{\|P\|_{\op}}{\sqrt{mn}} }   } \,.  
\end{equation}
Combining all estimates \eqref{As-A}, \eqref{hatA-As} on $\|\hat{A} - A_s\|_{2,\infty},  \|A_s-A\|_{2,\infty} $, we finally obtain 
\begin{equation*}
\begin{split}
  \|\hat{A}-A\|_{2,\infty} 
  & = 
    O\inparen{
        \log^4(m+n) K r^{3/2} \mu (\sqrt{m}+\sqrt{n}) \cdot 
        \insquare{
            \sqrt{\frac{r_p}{mq} + \frac{r_p}{nq} }+ \frac{\|P\|_{\op}}{\sqrt{mn}} 
        }  
    }.
\end{split}
  \end{equation*}

\subsection{High-level Proof Sketch of \cref{theo: mainRec}}

\label{sec:discussionThmmainRec}
To prove \cref{theo: mainRec}, we utilize a carefully designed adaptation of the contour integral approach that has been explored heavily in many recent works; \eg{}, \cite{ltranvufast, DKTranVu, tran2025davis, KX1, OVK22}. 
Instead of analyzing $\tilde{A}_s-A_s$ directly, we work with their symmetrized versions: 
\[
\mathcal{A} \coloneqq
\begin{bmatrix}0 & A \\ A^\top & 0\end{bmatrix},
\qquad
\mathcal{E} \coloneqq
\begin{bmatrix}0 & E \\ E^\top & 0\end{bmatrix},
\qquad
\tilde{\mathcal{A}} \coloneqq
\begin{bmatrix}0 & \tilde A \\ \tilde A^\top & 0\end{bmatrix}\,.
\]
The symmetrized versions are related to $\tilde{A}_s - A_s$ by
$$\textstyle \|\tilde{A}_s - A_s\|_{2,\infty} \coloneqq  \max_{1 \leq k \leq n} \|e_{n,k}^\top (\tilde{A}_s - A_s)\|_2 = \max_{1 \leq k \leq n} \|e_{m+n,k}^\top (\tilde{\mathcal{A}}_{2s} - \mathcal{A}_{2s})\big\|_2.$$
Thus, to prove \cref{theo: mainRec} it suffices to bound $\|e_1^\top (\tilde{\mathcal{A}}_{2s} - \mathcal{A}_{2s}) \|_2.$
Next, using the Cauchy integral theorem (\cref{theo: Cauchy}), we obtain
\[ 
\textstyle e_1^\top (\tilde{\mathcal{A}}_{2s} - \mathcal{A}_{2s}) = \frac{1}{2 \pi {\bf i} } \int_{\Gamma}  z \cdot e_1^\top  [(zI-\tilde{\mathcal{A}})^{-1})^{-1} - (zI- \mathcal{A})^{-1} ]  \, \d z, 
\]
where $\Gamma$ is a contour in $\mathbb{C}$ that encloses $\pm\sigma_1, \pm\sigma_2, \dots, \pm\sigma_s$ and excludes $ \pm \sigma_{s+1}, \pm\sigma_{s+2}, \dots, \pm\sigma_r$. 

\paragraph{Challenge (Prior techniques are insufficient to bound $\norm{\cdot}_{2,\infty}$ norm).}
If instead of the $\norm{\cdot}_{2,\infty}$, we wanted to bound the $\norm{\cdot}_2$ or $\norm{\cdot}_{\rm op}$ norms, then one could straightforwardly bound $\| e_1^\top  (\smash{\tilde{\mathcal{A}}}_{2s} - \mathcal{A}_{2s} )\|_2$ by $\| \smash{\tilde{\mathcal{A}}}_{2s} - \mathcal{A}_{2s} \|_{\op}$, and then apply the existing spectral-norm bounds on perturbations of low-rank approximations (\eg{}, \cite{TranVishnoiVu2025, tran2025newlowrank, EY1}). 
However, to obtain meaningful bounds on the Heterogenous treatment effect, we need to focus on the $\norm{\cdot}_{2,\infty}$ norm and, here, using the aforementioned bounds yields a suboptimal bound which can off by a large factor, of up to $\sqrt{n+m}$. 
Thus, obtaining a sharp bound with respect to the row-wise $\ell_2$-norm, which is crucial to prove \cref{theo: mainRec}, remains \mbox{a formidable analytical challenge that requires some new ideas.}

\paragraph{Ideas.}
To obtain the sharp bound, we carefully adapt the contour expansion method, introduced in \cite{DKTranVu, ltranvufast, TranVishnoiVu2025}. In particular, we repeatedly apply the Sherman--Morrison--Woodbury formula $\textstyle M^{-1} - (M+N)^{-1}= (M+N)^{-1} N M^{-1}$ \citep{HJBook} and $\tilde A=A+E$, to get
$$\textstyle e_1^\top (\tilde{\mathcal{A}}_{2s} - \mathcal{A}_{2s}) = \sum_{k=1}^\infty e_1^\top H_k ,\quadwhere\,H_k \coloneqq   \frac{1}{2 \pi \textbf{i}} \int_{\Gamma}  z \cdot (zI- \mathcal{A})^{-1} [ \mathcal{E}  (zI- \mathcal{A})^{-1}]^k \, \d z.$$
Intuitively, \cite{TranVishnoiVu2025} used a contour bootstrapping argument to show that
$\|\tilde{\mathcal A}_{2s}-\mathcal A_{2s}\|_{\op}$ is of the same order as $\|H_1\|_{\op}$. 
For entrywise control, \cite{ltranvufast} bounded $\|H_k\|_\infty$ for all $k\ge1$. 
In our row-wise $\ell_2$ setting, we instead bound $\|e_1^\top H_k\|_2$ for
$1\le k\le \ell\log n$, with a suitable constant $\ell$, and show that the remaining tail is negligible.

\paragraph{Further challenges.}
Bounding $\|e_1^\top H_k\|_2$ presents several further challenges. 
Sharp estimates require controlling the interactions
\(
\big|e_1^\top \mathcal E^\ell \binom{U}{V}\big|,
\forall 1\le \ell\le k.
\)
Naive bounds based on
$\|e_1^\top \mathcal E^\ell\|_2$, $\|\mathcal E\|_{\op}^\ell$, or
\(
\big\|\mathcal E^\ell\binom{U}{V}\big\|_2
\)
are suboptimal.  Since $\mathcal E$ is random, one expects $\mathcal E^\ell$ to spread mass across the entries of
$\binom{U}{V}^{\!\top}$, making the incoherence parameter $\mu$ essential. In \cite{ltranvufast}, the authors handled $\big|e_1^\top \mathcal E^\ell\binom{U}{V}\big|$ when $\mathcal{E}$ is random and mean-zero. In our setting, $\mathcal{E}=\mathcal{E}_R+\mathcal{E}_0$ contains both random and deterministic components, so separating their contributions is highly nontrivial; see \cref{sec: prooflemma}.
 
\section{Conclusion}

In this work, we study heterogeneous treatment-effect estimation in panel experiments where each unit is randomized many times, the propensities are unknown and may vary across units and time, and the potential-outcome matrices are approximately low-rank. 
We propose a simple row-scaled truncated-SVD estimator that uses only the observed outcomes and the assignment matrix, and we show that it recovers each unit's treatment-effect trajectory in a row-wise $\ell_2$ sense. 
This is enough to estimate treatment-effect averages for any individual unit over reasonably large subsets of times, which is the kind of guarantee one needs for the heterogeneous-effect questions that motivated the problem. 
At the heart of the analysis is a new perturbation bound for truncated SVD in the $\norm{\cdot}_{2,\infty}$ norm, which also makes precise how nonuniformity in the design affects estimation. Many natural questions remain. 
It would be interesting to relax the low-rank, incoherence, and signal-to-noise conditions, and to handle adaptive or dependent assignment mechanisms that arise in sequential experiments. 
Finally, sharpening the dependence on the design-nonuniformity term, especially in highly heterogeneous designs, is an interesting direction as well. %

\printbibliography
\newpage
\appendix

\newpage

\newpage

\section{Additional Discussion and Preliminaries}
In this section, we collect remarks on regularity conditions and some additional preliminaries. %
\subsection{Additional Remarks on \cref{asmp:main,asmp:main-formal}}
\label{sec:additional-remarks}
\begin{remark}[Thresholds]
\label{rem:thresholds}
In the theorem below we take $\tau_a=96KT(a)$ for $a\in\zo$.
Any threshold of the same order gives the same bound after changing
constants.  The truncation by $m^{-1}$ in the definition of $\wh p_i(a)$ is
only to avoid division by zero; under \cref{asmp:main-formal}, it is inactive
with high probability.
\end{remark}

\begin{remark}[Why the empirical rank is well-defined]
\label{rem:gap-exists}
The signal-to-noise condition ensures that the gap-selection step in
\cref{alg:row-scaled-svd} is nonempty with high probability.  Since
$\rank(A(a))\leq r$, we have
$\sigma_1(a)=\sum_{\ell=1}^r(\sigma_\ell(a)-\sigma_{\ell+1}(a))$, with
$\sigma_{r+1}(a)=0$.  Hence
$\sigma_1(a)>120KrT(a)$ implies that
$\sigma_s(a)-\sigma_{s+1}(a)>120KT(a)$ for some $s\leq r$.  On the
high-probability event
$\snorm{\wt Y^{\mathrm{ub}}(a)-A(a)}_{\op}\leq 12KT(a)$, Weyl's inequality
gives
$\wt\sigma^{\mathrm{ub}}_s(a)-\wt\sigma^{\mathrm{ub}}_{s+1}(a)\ge
96KT(a)$, so the algorithm selects at least one admissible truncation
level.  The selected $\wh s(a)$ is therefore an empirical effective rank,
not necessarily the algebraic rank of $A(a)$.
\end{remark}

\subsection{Some Classical Results} \label{others}

In this section, we recall standard results used in \cref{sec: main result}, \cref{sec: tech overview}, and \cref{sec: proofmainRec}.

\begin{theorem}[\textbf{Weyl's inequality}~\cite{We1}] \label{theo: weyl}
Let \( A, E\) be $n \times m$ matrices, and define \( \tilde{A}  \coloneqq  A + E \). Then, for any \( 1 \le i \le \min\{m,n\} \),
\[
|\tilde{\lambda}_i - \lambda_i| \le \|E\|_{\op} \quad \text{and} \quad |\tilde{\sigma}_i - \sigma_i| \le \|E\|_{\op},
\]
where \( \lambda_i, \tilde{\lambda}_i \) are the $i$th eigenvalues of \( A \) and \( \tilde{A} \), and \( \sigma_i, \tilde{\sigma}_i \) are the corresponding singular values.
\end{theorem}

\begin{theorem}[\textbf{Cauchy's integral theorem}] \label{theo: Cauchy}
Let $\Gamma$ be a simple closed contour, and let $f$ be an analytic function in the whole simply connected domain $S$ containing $\Gamma$. Then
$$
   \frac{1}{2 \pi {\bf i}} \int_{\Gamma} \frac{f(z)}{z-a}\,\, \d z 
   = 
   \begin{cases}
      f(a), & a \text{ inside } \Gamma, \\[4pt]
      0, & a \text{ outside } \Gamma
   \end{cases}.
$$
In particular, for $f(z)=z$, one has
$$
   \frac{1}{2 \pi {\bf i}} \int_{\Gamma} \frac{z}{z-a}\,\, \d z 
   = 
   \begin{cases}
      a, & a \text{ inside } \Gamma, \\[4pt]
      0, & a \text{ outside } \Gamma
   \end{cases}.
$$
\end{theorem}

\begin{theorem}[\textbf{Eckart--Young--Mirsky bound}~\cite{EY1}] \label{EY}
Let \( A, \tilde{A} \in \mathbb{R}^{n \times m} \), and let \( A_s\), \( \tilde{A}_s \) denote their respective best rank-$s$ approximations. Set \( E  \coloneqq  \tilde{A} - A \). Then,
\[
\| \tilde{A}_s - A_s \|_{\op} \le 2\left( \sigma_{s+1} + \| E \|_{\op} \right),
\]
where \( \sigma_{s+1} \) is the $(s+1)$th singular value of \( A \).
\end{theorem}

\section{Perturbation of Low-Rank Approximations in Row-Wise $\ell_2$ Norm} \label{sec: lowrankresult}
Let us recall the formal definition of row-wise $\ell_2$ norm. For each given pair of natural numbers $k \leq N$, denote the standard basis of $\mathbb{R}^N$ by 
$\{ e_{N,k} \,\,\text{for} \,\,1\leq k \leq N\}.$
Given an $n \times m$ matrix $M$ with $n$ rows $r_1, r_2,\cdots, r_n$. Define 
$$\|M\|_{2,\infty}: = \max_{1 \leq i \leq n} \|r_i\|_2 = \max_{1 \leq i \leq n} \|e_{n,i}^\top M\|_2\,.$$
Back to our setting of perturbation of low-rank approximations. Let $A$ be $n \times m$ matrix, admitting the singular decomposition:
$$A= \sum_{i=1}^r \sigma_i u_i v_i^\top\,,$$
where $\sigma_1 \geq \sigma_2 \geq \cdots \geq \sigma_r$ are singular values with the corresponding pair of left/right singular vectors $(u_i, v_i)_{1 \leq i \leq r}$. For each $1 \leq i \leq r-1$, the $i$th singular gap is $\delta_i \coloneqq  \sigma_i - \sigma_{i+1}$.

Let $E_R$ be an $n \times m$ random matrix of mean zero and $E_0$ be an $n \times m$ deterministic matrix. We denote 
$$E=E_R+E_0 \qquadand  \tilde{A}=A+E.$$ 
For a given rank-parameter $s$, our goal is to bound 
$$\|\tilde{A}_s -A_s\|_{2,\infty}.$$
Before stating our main bound, we recall the following notions and definitions. 

\paragraph{Incoherence number.} Define
$$\mu_R \coloneqq  \sqrt{n} \max_{1\leq i \leq n}\|e_{n,i}^\top U\|_2\qquadand \mu_C \coloneqq \sqrt{m} \cdot \max_{1 \leq j \leq m} \|V e_{m,j}\|_2.$$
The \textit{the incoherence number} of $A$ is $\mu=\max\{\mu_R, \mu_C\}$.

\begin{definition}
    A random matrix $E_R$ is called $(K,\sigma)$-bounded if $E_R$ has entries satisfying 
    $$\Ex E_{R,ij} =0,\, \Ex[|E_{R,ij}|^2] \leq \sigma^2,\, \Ex [|E_{R,ij}|^l] \leq K^{l-2} \sigma^2 \,\,\,\text{for all $l \geq 2 \in \mathbb{N}$ and $i \in [n], j \in [m]$.}$$
\end{definition}
We restate our key result - \cref{theo: mainRec}:

\paragraph{\cref{theo: mainRec}.} Let $E_R$ be a $(K,\sigma)$-bounded random matrix. There is a universal constant $C >0$ such that the following holds. If $\delta_s \geq 6(\|E_R\|_{\op}+\|E_0\|_{\op})$, then with probability at least $1-O(\nfrac{1}{(m+n)})$, 
    \begin{equation}
  \textstyle \frac{\|\tilde{A}_s-A_s\|_{2,\infty}}{C\sqrt{r}} \leq     \bigg[\log^2(m+n)+ \frac{K \log^4(m+n)}{\sqrt{m+n}} \bigg] \left(\frac{\mu}{\sqrt{m}}+ \frac{\mu}{\sqrt{n}}\right ) \cdot   \sqrt{m+n} \frac{\sigma_s}{\delta_s} \bigg[\sigma+\|E_0\|_{\op} \bigg].     
    \end{equation}
To ease presentation, let us now focus on the case that $n = \Theta(m)$, in which the RHS simplifies to 
$$ \bigg[\log^2 m+ \frac{K \log^4 m}{\sqrt{m}} \bigg] \big(\sigma+\|E_0\|_{\op} \big)  \mu \frac{\sigma_s}{\delta_s}.$$
We obtain the following corollaries in this setting. The first one is when $E_0 = 0$ (purely random noise), and the second one is when $\|E_0\|$ is large.  
\begin{corollary}[Random perturbation] \label{cor: random} 
    Under the above setting, let $E_R$ be a $(K,\sigma)$-bounded random matrix. There is a universal constant $C >0$ such that the following holds. If $\delta_s \geq 4\|E_R\|_{\op}$, then with probability at least $1-O(m^{-1})$, 
    \begin{equation}
   \|(A+E_R)_s-A_s\|_{2,\infty} \leq  C\sqrt{r}  \big[\log^2 m+ \frac{K \log^4 m}{\sqrt{m}} \big]  \mu \sigma \frac{\sigma_s}{\delta_s}.     
    \end{equation}
\end{corollary}
If $r, K, \sigma, \mu$ are all $\tilde{O}(1)$, then RHS becomes $\tilde{O}\big(\frac{\sigma_s}{\delta_s}\big)$. Moreover, in many widely studied structured models (\eg{}, spiked covariance, stochastic block, and graph Laplacian models), one typically has \(\sigma_{s} = O(\delta_{s})\), yielding the clean bound \(\tilde{O}(1)\). In comparison to the existing bounds on $\|(A+E_R)_s -A_s\|$ \eg{}, \cite{tran2025newlowrank, TranVishnoiVu2025, MangoubiVJACM, EY1}, which are at best $O(\|E_R\|)=O(\sqrt{m})$, our $\|\cdot\|_{2,\infty}$-bound is smaller by a factor $\sqrt{m}$. Since the optimal bound for $\|(A+E_R)_s-A_s\|_{\op}$ is $O(\|E_R\|_{\op})$, and 
$$\|(A+E_R)_s -A_s\|_{\op} \leq \sqrt{n} \|(A+E_R)_s-A_s\|_{2,\infty} = \Theta(\sqrt{m} \|(A+E_R)_s-A_s\|_{2,\infty})\,,$$
our bound is sharp up to a logarithmic factor.

\begin{corollary} Under the above setting, let $E_R$ be a $(K,\sigma)$-bounded random matrix and $\|E_0\| \gg \sigma$. There is a universal constant $C >0$ such that the following holds. If $\delta_s \geq 4(\|E_R\|_{\op}+\|E_0\|_{\op})$, then with probability at least $1-O(m^{-1})$, 
    \begin{equation}
   \|(A+E_R+E_0)_s-A_s\|_{2,\infty} \leq  C\sqrt{r}  \big[\log^2 m+ \frac{K \log^4 m}{\sqrt{m}} \big]  \mu \frac{\sigma_s}{\delta_s} \|E_0\|_{\op} .     
  \end{equation}
\end{corollary}
Similar to the discussion after Corollary \ref{cor: random}, in many typical settings the right-hand side simplifies to $\tilde O(\|E_0\|_{\op})$, which is sharp up to logarithmic factors. 
For example, let $m=n$, $E_R=0$, $A$ be diagonal, and $E_0=cI_m$. 
Then
\[
\|(A+E_R+E_0)_s - A_s\|_{2,\infty}
= \sigma_1 + c - \sigma_1
= c
= \|E_0\|_{\op}.
\]
Unlike random noise $E_R$, whose effect spreads across all directions, the deterministic component $E_0$ can have a localized impact on the perturbation.

\subsection{Proof of \cref{theo: mainRec}} \label{sec: proofmainRec}
In this subsection, we present the full detailed proof of \cref{theo: mainRec}. Recall the definition that for a given vector $\mathbf{u} \in \mathbb{R}^N,$  
$\|\mathbf{u}\|_\infty \coloneqq  \max_{1 \leq i \leq N} |\mathbf{u}_i|.$

\vskip2mm
\paragraph{Step 1 - Symmetrization.} We symmetrize $A$ and $\tilde{A}$ as follows. Set  $\mathcal{A} \coloneqq  \begin{pmatrix}
0 & A \\
A^\top & 0
\end{pmatrix}$, $ \mathcal{E} \coloneqq  \begin{pmatrix}
0 & E \\
E^\top & 0
\end{pmatrix},$ and then $\tilde{\mathcal{A}} = \begin{pmatrix}
    0 & \tilde{A} \\
    \tilde{A}^\top & 0
\end{pmatrix}.$
Given the singular decomposition of $A = U \Sigma V^\top$, it is easy to see  that $\mathcal{A}$ admits the following spectral decomposition
$$\mathcal{A}= \begin{pmatrix}
    \frac{U}{\sqrt{2}} & \frac{U}{\sqrt{2}} \\
    \frac{V}{\sqrt{2}} & -\frac{V}{\sqrt{2}}
\end{pmatrix} \begin{pmatrix}
    \Sigma & 0 \\
    0 & -\Sigma
\end{pmatrix} \begin{pmatrix}
    \frac{U^\top}{\sqrt{2}} & \frac{V^\top}{\sqrt{2}} \\
    \frac{U^\top}{\sqrt{2}} & - \frac{V^\top}{\sqrt{2}}
\end{pmatrix}\,,$$ %
that is,
$\mathcal{A}$ has the eigenvalues $\pm \sigma_i$ with the corresponding eigenvector $\mathbf{u}_{\pm i} \coloneqq  \frac{1}{\sqrt{2}} \begin{pmatrix}
     u_i \\
     \pm v_i
\end{pmatrix}.$ Similarly, $\tilde{\mathcal{A}}$ has the eigenvalues $\pm \tilde{\sigma}_i$ with the corresponding eigenvector $\tilde{\mathbf{u}}_{\pm i}$. 

The best rank-$2s$ approximation of $\mathcal{A}$ is
$$\mathcal{A}_{2s} \coloneqq \begin{pmatrix}
    \frac{U}{\sqrt{2}} & \frac{U}{\sqrt{2}} \\
    \frac{V}{\sqrt{2}} & -\frac{V}{\sqrt{2}}
\end{pmatrix} \begin{pmatrix}
    \Sigma_s & 0 \\
    0 & -\Sigma_s
\end{pmatrix} \begin{pmatrix}
    \frac{U^\top}{\sqrt{2}} & \frac{V^\top}{\sqrt{2}} \\
    \frac{U^\top}{\sqrt{2}} & - \frac{V^\top}{\sqrt{2}}
\end{pmatrix}, $$
which, in fact, equals 
$$\mathcal{A}_{2s} = \begin{pmatrix}
     0 & U \Sigma_s V^\top \\
    V \Sigma_s U^\top & 0
\end{pmatrix}.$$
Similarly,
$$\tilde{\mathcal{A}}_{2s} = \begin{pmatrix}
     0 & \tilde U \tilde\Sigma_s \tilde V^\top \\
    \tilde V \tilde \Sigma_s \tilde U^\top & 0
\end{pmatrix},$$
and then,  
$$\tilde{\mathcal{A}}_{2s} - \mathcal{A}_{2s} = \begin{pmatrix}
     0 & \tilde{U} \tilde{\Sigma}_s \tilde{V}^\top- U \Sigma_s V^\top \\
   \tilde{V} \tilde{\Sigma}_s \tilde{U}^\top - V \Sigma_s U^\top & 0
\end{pmatrix}.$$
It yields
 $$ \big\|e_1^\top(\tilde{A}_s -A_s)\big\|_2=\big\| e_1^\top (\tilde{U} \tilde{\Sigma}_s \tilde{V}^\top- U \Sigma_s V^\top)\big\|_2 = \big\|e_1^\top (\tilde{\mathcal{A}}_{2s} - \mathcal{A}_{2s})\big\|_2,$$  
and then
$$\|\tilde{A}_s - A_s\|_{2,\infty} \coloneqq  \max_{1 \leq k \leq n} \|e_{n,k}^\top (\tilde{A}_s - A_s)\|_{2} = \max_{1 \leq k \leq n} \|e_{m+n,k}^\top (\tilde{\mathcal{A}}_{2s} - \mathcal{A}_{2s})\big\|_2.$$

\vskip2mm

\paragraph{Step 2 - Contour representation of perturbation.} Let $\Gamma$ be a contour in $\mathbb{C}$ that encloses $\pm\sigma_1, \pm\sigma_2, \dots, \pm\sigma_s$ and excludes $ \pm \sigma_{s+1}, \pm\sigma_{s+2}, \dots, \pm\sigma_r$. The well-known contour integral representation \cite{Higham2008, Kato1, SS1} gives us:
\[ \frac{1}{2 \pi \textbf{i}} \int_\Gamma z (zI -\mathcal{A})^{-1} \, \d z =  \sum_{i  =1}^s \sigma_i  \mathbf{u}_{+i} \mathbf{u}_{+i}^\top + \sum_{i=1}^s (-\sigma_i) \mathbf{u}_{-i}  \mathbf{u}_{-i}^\top = \mathcal{A}_{2s}.\] 
Let $\tilde{\lambda}_1 \geq \cdots \geq \tilde{\lambda}_n$ denote the eigenvalue of $\tilde{A}$ with the corresponding eigenvectors $\tilde{u}_1, \tilde{u}_2, \dots, \tilde{u}_n$. The construction of $\Gamma$ (presented later) and the gap assumption $4\|E\| < \delta_s$ ensure that the eigenvalues \( \tilde{\lambda}_i \)  lie inside \( \Gamma \) if and only if $|\tilde\lambda_i| \in \{\pm \tilde{\sigma}_1, \pm \tilde{\sigma}_2, \dots, \pm \tilde{\sigma}_s  \}$.  Then, similarly, we have
  \(\textstyle \frac{1}{2 \pi {\bf i}}  \int_{\Gamma}  z (z I-\tilde{\mathcal{A}})^{-1} \, \d z  = \sum_{i  =1}^s \tilde\sigma_i  \tilde{\mathbf{u}}_{+i} \tilde{\mathbf{u}}_{+i}^\top + \sum_{i=1}^s (- \tilde{\sigma}_i) \tilde{\mathbf{u}}_{-i}  \tilde{\mathbf{u}}_{-i}^\top = \tilde{\mathcal{A}}_{2s} .\)

\noindent Thus, we obtain the following contour identity for the perturbation: 
\begin{equation} \label{contourRepres}
     \tilde{\mathcal{A}}_{2s} - \mathcal{A}_{2s} =  \frac{1}{2 \pi {\bf i} } \int_{\Gamma}  z [(zI-\tilde{\mathcal{A}})^{-1})^{-1} - (zI- \mathcal{A})^{-1} ] \, \, \d z.  
\end{equation}
Therefore, 
$$ e_1^\top \left(\tilde{\mathcal{A}}_{2s} - \mathcal{A}_{2s} \right) = \frac{1}{2 \pi {\bf i} } \int_{\Gamma}  z \cdot e_1^\top  [(zI-\tilde{\mathcal{A}})^{-1})^{-1} - (zI- \mathcal{A})^{-1} ]  \, \, \d z. $$
This contour method is widely used in functional perturbation analysis, \eg{}, \cite{Higham2008, Kato1, KX1, OVK22, tran2025davis, TranVishnoiVu2025, DKTranVu, TranVishnoi2025}. Note that one can trivially bound $\| e_1^\top  \left(\tilde{\mathcal{A}}_{2s} - \mathcal{A}_{2s} \right)\|_2$ by $\| \tilde{\mathcal{A}}_{2s} - \mathcal{A}_{2s} \|_{\op}$, and then apply the existing bounds (\eg{}, \cite{TranVishnoiVu2025, tran2025newlowrank, EY1}) in perturbation theory. However, this approach only gives a suboptimal bound that is off by a factor of up to $\sqrt{n+m}$. Thus, obtaining a sharp bound with respect to the $\ell_1$-norm remains a formidable analytical challenge.

\vskip2mm

\paragraph{Step 3 - Contour expansion.} We adapt the contour expansion method, introduced in \cite{DKTranVu, ltranvufast, TranVishnoiVu2025}. Using the Sherman-Morrison-Woodbury formula $\textstyle M^{-1} - (M+N)^{-1}= (M+N)^{-1} N M^{-1}$ \citep{HJBook} and the fact that $\textstyle \tilde A =A+E$, we obtain 
\[\textstyle (zI- \mathcal{A})^{-1} - (zI-\tilde {\mathcal{A}})^{-1} =  (zI-\mathcal{A})^{-1} \mathcal{E} (zI-\tilde{\mathcal{A}})^{-1},
\]
and hence 
\[ \textstyle (zI- \mathcal{A})^{-1} - (zI-\tilde {\mathcal{A}})^{-1} = (zI-\mathcal{A})^{-1} \mathcal{E}  (zI-\mathcal{A})^{-1} + (zI-\mathcal{A})^{-1} \mathcal{E}  [(zI-\tilde{\mathcal{A}})^{-1} - (zI- \mathcal{A})^{-1}].
\]
By repeatedly applying the resolvent formula $L$ times, one can expand $z e_1^\top [ (zI -\tilde{A})^{-1} -(zI -A)^{-1}]$ into
\[ \big(\sum_{k=1 }^{L} z \cdot e_1^\top (zI- \mathcal{A})^{-1} [ \mathcal{E}  (zI- \mathcal{A})^{-1}]^k \big) + e_1^\top (zI- \mathcal{A})^{-1} [ \mathcal{E}  (zI- \mathcal{A})^{-1}]^{L-1} \cdot [(zI-\tilde{\mathcal{A}})^{-1} - (zI- \mathcal{A})^{-1}].
\]
Therefore, 
\begin{equation}
    \begin{split}
        & e_1^\top \left(\tilde{\mathcal{A}}_{2s} - \mathcal{A}_{2s} \right)  = \left(\sum_{k=1}^L  F_k \right) + F_{\mathrm{tail}},\,\,\,\text{where} \\
        & F_k \coloneqq    \frac{1}{2 \pi \textbf{i}} \int_{\Gamma}  z \cdot e_1^\top (zI- \mathcal{A})^{-1} [ \mathcal{E}  (zI- \mathcal{A})^{-1}]^k \, \d z   \quad \text{and} \\
        & F_{\mathrm{tail}} \coloneqq   \frac{1}{2 \pi \textbf{i}} \int_{\Gamma}  e_1^\top (zI- \mathcal{A})^{-1} [ \mathcal{E}  (zI- \mathcal{A})^{-1}]^{L-1} \cdot [(zI-\tilde{\mathcal{A}})^{-1} - (zI- \mathcal{A})^{-1}] \, \d z .
    \end{split}
\end{equation}
Thus, by the triangle inequality, we have
$$\big\|e_1^\top \left(\tilde{\mathcal{A}}_{2s} - \mathcal{A}_{2s} \right) \big\|_2 \leq \left(\sum_{k=1}^L \| F_k\|_2 \right) + \|F_{\mathrm{tail}}\|_2.$$ 
We set $L=10 \ell \log (m+n)$ (the constant $\ell$ will be chosen later). The remaining duty is to bound $\|F_k\|_2$ for each $1 \leq k \leq L$ and $\|F_{\mathrm{tail}}\|_2$. Indeed, we are going to show that the main part is $\sum_{k=1}^L \|F_k\|_2$ and $\|F_{\mathrm{tail}}\|_2$ is negligible.

Inspired by the construction of $\Gamma$ in \cite{tran2025davis, TranVishnoiVu2025, tran2026eigenvalue}, we set $\Gamma$ as as the union of two disjoint rectangles $\Gamma^{+} \cup \Gamma^{-}$, where:  
\begin{itemize}
    \item $\Gamma^{+}$ isolates $\{\sigma_1, \sigma_2, \dots, \sigma_s\}$, with (1) its left vertical edge intersecting the real axis at $a_0 \coloneqq \sigma_s - \delta_s/2$; (2) its right vertical edge intersecting the real axis at $a_1 \coloneqq 2\sigma_1$; (3) its height (from the real line) is $T=2\sigma_1$.
    \item $\Gamma^{-}$ isolates $\{-\sigma_1, -\sigma_2, \dots,- \sigma_s\}$, with (1) its left vertical edge intersecting the real axis at $b_1 \coloneqq -2\sigma_1$; (2) its right vertical edge intersecting the real axis at $b_0 \coloneqq -\sigma_s+\delta_s/2$; (3) its height (from the real line) is $T=2\sigma_1$.
\end{itemize}
See the figure below for an illustration.

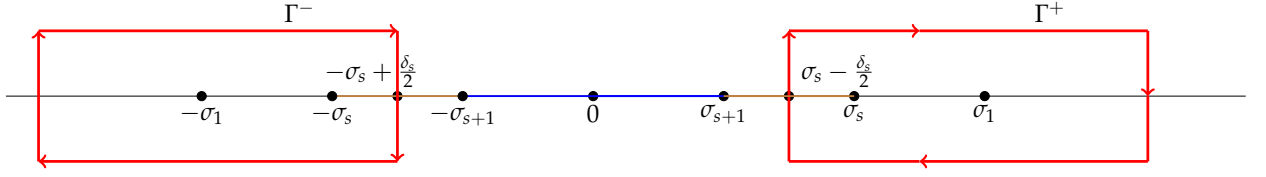
\resizebox{\linewidth}{!}{
\begin{tikzpicture}
\coordinate (A) at (9,0);
\node[below] at (A){$0$};
\coordinate (A') at (11, 0);
\node[below] at (A'){$\sigma_{s+1}$};
\coordinate (C) at (13,0);
\node[below] at (C){$\sigma_{s}$};
\coordinate (D) at (17.5,0);
\coordinate (S1) at (12.75,0);
\node[above] at (S1){$\sigma_s-\frac{\delta_s}{2}$};
\coordinate (A1) at (15,0);
\node[below] at (A1){$\sigma_{1}$};
\coordinate (S2) at (5.6,0);
\node[above] at (S2){$-\sigma_s+\frac{\delta_s}{2}$};
\coordinate (G1) at (4.5,1);
\node[above] at (G1){$\Gamma^{-}$};
\coordinate (G2) at (16,1);
\node[above] at (G2){$\Gamma^{+}$};
\coordinate (B) at (12,0);

\coordinate (B') at (12,1);
\coordinate (F) at (14,1);

\coordinate (G) at (14,-1);

\draw[-] (0,0) -- (19,0);

\draw[thick,blue] (A) -- (A'); %
\draw[thick,brown] (A')  -- (C); %

\filldraw (A1) circle (2pt);
\filldraw (A) circle (2pt);
\filldraw (A') circle (2pt);
\filldraw (C) circle (2pt);
\filldraw (B) circle (2pt);

\draw[very thick,red,->] (12,1) -- (14,1); %
\draw[very thick,red] (14,1) -- (17.5,1); 
\draw[very thick,red,->] (17.5,1) -- (D); %
\draw[very thick,red] (D) -- (17.5,-1); %
\draw[very thick,red,->] (17.5,-1) -- (14,-1); %
\draw[very thick,red] (14,-1) -- (12,-1); 
\draw[very thick,red,->] (12,-1) -- (B'); %
\draw[very thick,red] (B') -- (12,1); %

\coordinate (A'1) at (7, 0);
\node[below] at (A'1){$-\sigma_{s+1}$};
\coordinate (C'1) at (5,0);
\node[below] at (C'1){$-\sigma_{s}$};

\coordinate (D'1) at (0.5,0);
\coordinate (A11) at (3,0);
\node[below] at (A11){$-\sigma_{1}$};

\coordinate (B'1) at (6,0);
\coordinate (B'11) at (6,1);

\filldraw (B'1) circle (2pt);
\draw[thick,blue] (A) -- (A'); %
\draw[thick,brown] (A')  -- (C); %
\draw[thick,blue] (A) -- (A'1);
\draw[thick,brown] (A'1) -- (C'1);
\filldraw (A'1) circle (2pt);
\filldraw (C'1) circle (2pt);
\filldraw (A11) circle (2pt);
\draw[very thick,red,->] (0.5,1) -- (6,1); %
\draw[very thick,red,->] (6,1) -- (6,-1); %
\draw[very thick,red,->] (6,-1) -- (0.5,-1); %
\draw[very thick,red,->] (0.5,-1) -- (0.5,1); %
\end{tikzpicture}}

\vskip2mm

\paragraph{Step 4 - Bounding $\|F_k\|$ for a natural number $k \leq L$.} We split the spectral decomposition of $(zI -\mathcal{A})^{-1} = \sum_{i=1}^{r} \frac{\mathbf{u}_{+i}\mathbf{u}_{+i}^\top}{z-\sigma_i} + \sum_{i=1}^{r} \frac{\mathbf{u}_{-i}\mathbf{u}_{-i}^\top}{z+\sigma_i} + \frac{I_{m+n} - (\sum_{i=1}^r \mathbf{u}_{+i}\mathbf{u}_{+i}^\top+ \mathbf{u}_{-i}\mathbf{u}_{-i}^\top)  }{z}$ into $P+Q$, where
\begin{equation}
    \begin{split}
        P
        \quad&\coloneqq\quad  \sum_{i=1}^{r} \frac{\mathbf{u}_{+i}\mathbf{u}_{+i}^\top}{z-\sigma_i} + \sum_{i=1}^{r} \frac{\mathbf{u}_{-i}\mathbf{u}_{-i}^\top}{z+\sigma_i} - \frac{\sum_{i=1}^r \mathbf{u}_{+i}\mathbf{u}_{+i}^\top+ \mathbf{u}_{-i}\mathbf{u}_{-i}^\top}{z} \\
        & = \quad\sum_{i=1}^r \frac{\sigma_i}{(z-\sigma_i)z} \mathbf{u}_{+i}\mathbf{u}_{+i}^\top+ \sum_{i=1}^{r}\frac{-\sigma_i}{(z+\sigma_i)z} \mathbf{u}_{-i}\mathbf{u}_{-i}^\top,\\
        \text{and}&\quad Q = \frac{I_{m+n}}{z}.
    \end{split}
\end{equation}
We can rewrite $2 \pi \textbf{i} F_k$ as what follows 
\begin{equation}
    \begin{split}
     &      \int_{\Gamma}  z \cdot e_1^\top P [ \mathcal{E}  (zI- \mathcal{A})^{-1}]^k \, \d z +  \int_{\Gamma}  z \cdot e_1^\top Q [ \mathcal{E}  (zI- \mathcal{A})^{-1}]^k \, \d z \\
    & =   \int_{\Gamma}  z \cdot e_1^\top P [ \mathcal{E}  (zI- \mathcal{A})^{-1}]^k \, \d z +  \int_{\Gamma}  z \cdot \frac{1}{z} e_1^\top  \mathcal{E}  (zI- \mathcal{A})^{-1} \cdot [\mathcal{E}  (zI- \mathcal{A})^{-1}]^{k-1} \, \d z \\
    & = \int_{\Gamma}  z \cdot e_1^\top P [ \mathcal{E}  (zI- \mathcal{A})^{-1}]^k \, \d z +  \int_{\Gamma}  z \cdot \frac{1}{z} e_1^\top  \mathcal{E} P \cdot [\mathcal{E}  (zI- \mathcal{A})^{-1}]^{k-1} \, \d z +  \int_{\Gamma}  z \cdot \frac{1}{z^2} e_1^\top  \mathcal{E}   \cdot [\mathcal{E}  (zI- \mathcal{A})^{-1}]^{k-1} \, \d z \\
    & = \cdots \\
    & = \left[\sum_{l=0}^k  \int_{\Gamma}  z \cdot \frac{1}{z^{l}} e_1^\top \mathcal{E}^l P [ \mathcal{E}  (zI- \mathcal{A})^{-1}]^{k-l} \, \d z \right] +  \int_{\Gamma}  z \cdot \frac{1}{z^{k+1}} e_1^\top E^k \, \d z. 
    \end{split}
\end{equation}
Note that $z = 0$ is outside of the contour $\Gamma$. Thus, the last term $ \int_{\Gamma}  z \cdot \frac{1}{z^{k+1}} e_1^\top E^k \, \d z$ is zero, and hence 
$$F_k \coloneqq  \sum_{l=0}^k \frac{1}{2\pi \textbf{i}} \int_{\Gamma}  z \cdot \frac{1}{z^{l}} e_1^\top E^l P [ \mathcal{E}  (zI- \mathcal{A})^{-1}]^{k-l} \, \d z. $$
By the triangle inequality, we have
$$\|F_k\|_2 \leq \sum_{l=0}^k G_{k,l}\quadwhere G_{k,l} \coloneqq  \frac{1}{2\pi} \cdot \int_{\Gamma}  \big\| \frac{1}{z^{l-1}} e_1^\top \mathcal{E}^l P [ \mathcal{E}  (zI- \mathcal{A})^{-1}]^{k-l}\big\|_2\,|\, \d z|.$$
If $l \geq 1$, splitting out the factor $[ \mathcal{E}  (zI- \mathcal{A})^{-1}]^{k-l}$, we have 
$$G_{k,l} \leq \frac{\max_{z \in \Gamma} \| \mathcal{E}  (zI- \mathcal{A})^{-1}\|_{\op}^{k-l}}{2 \pi} \cdot \int_{\Gamma} \big\| \frac{1}{z^{l-1}} e_1^\top \mathcal{E}^l P\big\|_2 \,|\, \d z| \leq \frac{1}{2^{k-l+1} \pi} \cdot  \int_{\Gamma} \big\| \frac{1}{z^{l-1}} e_1^\top \mathcal{E}^l P\big\|_2 \,|\, \d z|.$$
The last inequality is true by our gap assumption that $4 \|\mathcal{E}\|_{\op} < \delta_s$ and the construction of $\Gamma$, which implies $\min_{z \in \Gamma, i \in [n]}|z -\lambda_i| \geq \delta_s/2 \geq 2 \|E\|$, and hence
\begin{equation*} \label{gapE}
    \max_{z \in \Gamma} \| \mathcal{E}  (zI- \mathcal{A})^{-1}\|_{\op} \leq \max_{z \in \Gamma} \| (zI -\mathcal{A})^{-1} \|_{\op} \cdot \|\mathcal{E}\|_{\op} = \frac{\|\mathcal{E}\|_{\op}}{\min_{z \in \Gamma, i \in [n]} |z \pm \sigma_i|} \leq  \frac{\|\mathcal{E}\|_{\op}}{2 \| \mathcal{E} \|_{\op} } = \frac{1}{2}.
\end{equation*}
For the second factor, we use the following lemma. Its proof is delayed to the next section. 

\vskip2mm
\begin{lemma} \label{lem: concenElke1}
  Given a natural number $l \geq 1$.  Under the assumption of \cref{theo: mainRec}, with probability at least $1 -O((m+n)^{-2})$, there is a universal constant $C > 0$ such that $  \frac{\int_{\Gamma} \big\| \frac{1}{z^{l-1}} e_1^\top \mathcal{E}^l P\big\|_2 \,|\, \d z|}{2\pi C \sqrt{r}} $ is at most
    \begin{equation*}
     \frac{1}{3^{l-1}}\cdot (\sigma \sqrt{m+n} \frac{\sigma_s}{\delta_s}) \left( \bigg[\log^2(m+n)+ \frac{K \log^4(m+n)}{\sqrt{m+n}} \bigg]\max_{1 \leq i \leq r} \|\mathbf{u}_{\pm i}\|_{\infty} + \frac{\|\mathcal{E}_0\|_{\op} \log (m+n)}{\sigma \sqrt{m+n}} \right).
        \end{equation*}
\end{lemma}
Thus, by Lemma \eqref{lem: concenElke1}, we further obtain that $G_{k,l}, l \geq 1$ is at most
\begin{equation}\label{Gkl>0}
    \frac{C\sqrt{r}}{2^k} \big(\frac{2}{3} \big)^{l-1} \cdot (\sigma \sqrt{m+n} \frac{\sigma_s}{\delta_s}) \left( \bigg[\log^2(m+n)+ \frac{K \log^4(m+n)}{\sqrt{m+n}} \bigg]\max_{1 \leq i \leq r} \|\mathbf{u}_{\pm i}\|_{\infty} + \frac{\|\mathcal{E}_0\|_{\op} \log (m+n)}{\sigma \sqrt{m+n}} \right),
\end{equation}
with probability at least $1 -O((m+n)^{-2})$ for some universal constant $C$.

\vskip2mm

Next, for $l =0$, splitting out the factor $[ \mathcal{E}  (zI- \mathcal{A})^{-1}]^{k-1}$, we have 
\begin{equation*}
   \begin{split}
     G_{k,0} & \leq \frac{\max_{z \in \Gamma} \| \mathcal{E}  (zI- \mathcal{A})^{-1}\|_{\op}^{k-1}}{2 \pi} \cdot \int_{\Gamma} \big\| z e_1^\top P \mathcal{E} (zI- \mathcal{A})^{-1}\big\|_2 \,|\, \d z| \\
     &\leq \frac{1}{2^{k-1} \pi} \cdot  \int_{\Gamma} \big\| z e_1^\top P \mathcal{E} (zI- \mathcal{A})^{-1}\big\|_2 \,|\, \d z|.  
   \end{split}
\end{equation*}
Similarly, for the second factor, we use the following technical lemma, whose proof will be presented in the next section. 

\vskip2mm
\begin{lemma} \label{lem: concenE0ke1}
   Under the assumption of \cref{theo: mainRec}, with probability at least $1 -O((m+n)^{-2})$, there is a universal constant $C > 0$ such that 
    $$\int_{\Gamma} \big\| z e_1^\top P \mathcal{E} (zI- \mathcal{A})^{-1}\big\|_2 \,|\, \d z| \leq C\sqrt{r} \log^2(m+n) \cdot (\|\mathcal{E}_0\|_{\op}+\sigma)  (\sqrt{m+n} \frac{\sigma_s}{\delta_s}) \cdot \max_{1 \leq i \leq r} \|\mathbf{u}_{\pm i}\|_{\infty}.$$
\end{lemma}
By \cref{lem: concenE0ke1}, with probability at least $1-O((m+n)^{-2})$, 
$$\int_{\Gamma} \big\| z e_1^\top P \mathcal{E} (zI- \mathcal{A})^{-1}\big\|_2 \,|\, \d z| \leq C\sqrt{r} \log^2(m+n) \cdot  (\|\mathcal{E}_0\|_{\op}+\sigma)  ( \sqrt{m+n} \frac{\sigma_s}{\delta_s}) \cdot \max_{1 \leq i \leq r} \|\mathbf{u}_{\pm i}\|_{\infty}.$$
Thus, 
\begin{equation} \label{Gk0}
    G_{k,0} \leq \frac{C\sqrt{r}}{2^k} \log^2(m+n) \cdot (\|\mathcal{E}_0\|_{\op}+\sigma) \cdot(\sqrt{m+n} \frac{\sigma_s}{\delta_s}) \cdot \max_{1 \leq i \leq r} \|\mathbf{u}_{\pm i}\|_{\infty}.
\end{equation}
The estimates on $G_{k,l}$ and the fact that $\max_{1 \leq i \leq r} \|\mathbf{u}_{\pm i}\|_{\infty} \geq \frac{1}{\sqrt{m+n}}$, imply that with probability at least $1 -O(k (m+n)^{-2}),$ there is a universal constant $C>0$, such that 
\begin{equation}
    \begin{split}
   \|F_k\| & \leq \sum_{l=0}^k G_{k,l} \\
   & \leq \frac{C\sqrt{r}}{2^k} \cdot (\|\mathcal{E}_0 \|_{\op}+\sigma) ( \sqrt{m+n} \frac{\sigma_s}{\delta_s})  \bigg[\log^2(m+n)+ \frac{K \log^4(m+n)}{\sqrt{m+n}} \bigg]\max_{1 \leq i \leq r} \|\mathbf{u}_{\pm i}\|_{\infty}.         \end{split}
\end{equation}

\paragraph{Step 5 - Bounding $\|F_{\mathrm{tail}}\|$.} We have
\begin{equation} \label{Ineq: Ftail0}
    \begin{split}
      \|F_{\mathrm{tail}}\|_2 & =   \frac{1}{2 \pi} \cdot \bigg\| \int_{\Gamma}  e_1^\top (zI- \mathcal{A})^{-1} [ \mathcal{E}  (zI- \mathcal{A})^{-1}]^{L-1} \cdot [(zI-\tilde{\mathcal{A}})^{-1} - (zI- \mathcal{A})^{-1}] \, \d z \bigg\|_2 \\
      & \leq \frac{10\sigma_1}{2 \pi} \cdot \max_{z \in \Gamma} \| (zI- \mathcal{A})^{-1} [ \mathcal{E}  (zI- \mathcal{A})^{-1}]^{L-1} \|_{\op} \cdot \left(\|(zI-\tilde{\mathcal{A}})^{-1}\|_{\op}+\|(zI- \mathcal{A})^{-1}\|_{\op} \right) \\
      & \leq \frac{10 \sigma_1}{2\pi} \cdot \|\mathcal{E}\|_{\op}^{L-1} \cdot \max_{z \in \Gamma} \|(zI- \mathcal{A})^{-1}\|_{\op}^L \cdot \left(\|(zI-\tilde{\mathcal{A}})^{-1}\|_{\op}+\|(zI- \mathcal{A})^{-1}\|_{\op} \right).
    \end{split}
    \end{equation}
The first inequality is obtained by the fact that the length of the contour $\Gamma$ is at most $10\sigma_1$. Moreover, 
$ \max_{z \in \Gamma} \|(zI- \mathcal{A})^{-1}\|_{\op} = \frac{1}{\min_j |z -\lambda_j|} = \frac{1}{\delta_s/2} = \frac{2}{\delta_s}, $ the RHS is at most 
\begin{equation} \label{ineq: Ftail1}
    \frac{10 \sigma_1}{\pi \delta_s} \cdot (2\|\mathcal{E}\|_{\op}/\delta_s) ^{L-1} \cdot \left(\|(zI-\tilde{\mathcal{A}})^{-1}\|_{\op}+ \frac{2}{\delta_s} \right).
\end{equation}
Next, we upper bound $\|(zI-\tilde{\mathcal{A}})^{-1}\|_{\op}$.   For $z = a+ \textbf{i} b,$ we have
$$\sigma_{\min}(zI-\tilde{\mathcal{A}}) = \min_{1 \leq j \leq m+n} \sqrt{b^2+ (a- \tilde{\lambda}_j)^2},$$
here $\tilde{\lambda}_j$ are eigenvalues of $\tilde{\mathcal{A}}.$ Let $\ell$ be the natural number such that $\sigma_1 \leq n^\ell$. We have the following cases. 
\begin{itemize}[leftmargin=15pt]
    \item \textbf{Case 1:} $b \geq (m+n)^{-\ell}$. Thus, $\sigma_{\min}(zI-\tilde{\mathcal{A}}) \geq (m+n)^{-\ell}$, and hence $\|(zI-\tilde{\mathcal{A}})^{-1}\|_{\op} \leq(m+n)^{\ell}.$
    \item \textbf{Case 2:} $b < n^{\ell}$. So $z$ must be on the vertical sides of $\Gamma$, \ie{} $\mathcal{R}e z$ is either $\pm 2\sigma_1$ or $\pm (\sigma_s -\delta_s/2)$ If $\mathcal{R}e z = 2\sigma_1$, then 
    $$\sigma_{\min}(zI-\tilde{\mathcal{A}}) \geq \sigma_1 - \|\mathcal{E}\|_{\op} \geq \sigma_1/2,\,\,\text{and hence}\,\,\|(zI-\tilde{\mathcal{A}})^{-1}\|_{\op} \leq \frac{2}{\sigma_1}.$$
    If $\mathcal{R}e z=\sigma_s -\delta_s/2,$ then
        $$\|zI - \mathcal{A}-\mathcal{E}_0\|_{\op} \leq n^{-\ell}+ \delta_s/2 +\|\mathcal{E}_0\|_{\op} \leq \delta_s \leq \sigma_1.$$
    By applying \citep{JSS1, tao2008random} on the pair of $(zI - \mathcal{A}-\mathcal{E}_0, zI - \tilde{\mathcal{A}}) $ with respect to the random noise $\mathcal{E}_R$, with probability at least $1- (m+n)^{-C_2}$
    $$\sigma_{\min}(zI-\tilde{\mathcal{A}}) \geq (m+n)^{-2(C_2+2) \ell+\frac{1}{2}+o(1)}.$$
It means, with probability at least $1-(m+n)^{-2}$, 
  $$\|(zI-\tilde{\mathcal{A}})^{-1}\|_{\op} \leq (m+n)^{8 \ell+\frac{1}{2}+o(1)}.$$
\end{itemize}
All cases imply that with probability at least $1-(m+n)^{-2},$
$$\|(zI-\tilde{\mathcal{A}})^{-1}\|_{\op} \leq (m+n)^{8\ell+\frac{1}{2}+o(1)}.$$

\noindent Combining \eqref{Ineq: Ftail0} with \eqref{ineq: Ftail1} and the upper bound of $\|(zI-\tilde{\mathcal{A}})^{-1}\|$, we obtain that with probability at least $1 -(m+n)^{-2},$
$$\|F_{\mathrm{tail}}\|_2 \leq \bigg[\frac{\|\mathcal{E}\|_{\op} \cdot \sigma_s}{\delta_s} \cdot \frac{1}{\sqrt{m+n}} \bigg] \cdot \frac{(m+n)^{9 \ell+1+o(1)}}{(\delta_s/2\|\mathcal{E}\|_{\op})^{L-1}} \leq  \bigg[\frac{\|\mathcal{E}\|_{\op} \cdot \sigma_s}{\delta_s} \cdot \frac{1}{\sqrt{m+n}} \bigg] \cdot \frac{(m+n)^{9\ell+1+o(1)}}{2^{L-1}}.$$
Thus, by setting $L= 10 \ell \cdot\log(m+n),$ we have 
$$\|F_{\mathrm{tail}}\|_2 \leq o  \bigg(\frac{\|\mathcal{E}\|_{\op} \cdot \sigma_s}{\delta_s} \cdot \frac{1}{\sqrt{m+n}} \bigg) ,$$
which is negligible in comparison to $\sum_{k=1}^L \|F_k\|_2.$

We finally obtain that 
$$\big\|e_1^\top \left(\tilde{\mathcal{A}}_{2s} - \mathcal{A}_{2s} \right) \big\|_2 \leq C\sqrt{r} \cdot (\|\mathcal{E}_0 \|_{\op}+\sigma) ( \sqrt{m+n} \frac{\sigma_s}{\delta_s})  \bigg[\log^2(m+n)+ \frac{K \log^4(m+n)}{\sqrt{m+n}} \bigg]\max_{1 \leq i \leq r} \|\mathbf{u}_{\pm i}\|_{\infty}.$$
This completes our proof. 
\begin{remark}
    We can slightly improve the upper bound on $\|F_k\|_2$ by combining exactly the upper bounds of $G_{k,0}, G_{k,l}$ without replacing $\frac{1}{\sqrt{m+n}}$ by $\max_{1 \leq i \leq r} \|\mathbf{u}_{\pm i}\|_{\infty}.$ Indeed, $\|F_k\|_2/(C\sqrt{r})$ is at most 
    $$\frac{\log^2(m+n)+ \frac{K \log^4(m+n)}{\sqrt{m+n}}}{2^k} \cdot ( \sqrt{m+n} \frac{\sigma_s}{\delta_s})  \bigg[ ( \mathcal{X}_0+\sigma) \cdot  \max_{1 \leq i \leq r} \|\mathbf{u}_{\pm i}\|_{\infty}+\frac{\|\mathcal{E}_0\|_{\op}}{\sqrt{m+n}} \bigg], $$
    where $\mathcal{X}_0 \coloneqq  \max_{1 \leq i \leq r, 1 \leq j \leq m+n} |\mathbf{u}_{\pm i}^\top \mathcal{E}_0 \mathbf{v}_j|$ with $\mathbf{v}_j, 1\leq j \leq m+n$ eigenvectors of $\mathcal{A}.$
    Define $X_0 \coloneqq  \max_{1\leq i,j \leq r} |u_i^\top E_0 v_j|$. As a result, we obtain that $\|\tilde{A}-A\|_{2,\infty} $ is at most, 
$$ \bigg[\log^2(m+n)+ \frac{K \log^4(m+n)}{\sqrt{m+n}}\bigg] \cdot ( \sqrt{m+n} \frac{\sigma_s}{\delta_s})  \bigg[ \mu \big(\frac{1}{\sqrt{m}}+\frac{1}{\sqrt{n}} \big)( X_0+\sigma) +\frac{\|E_0\|_{\op}}{\sqrt{m+n}} \bigg],$$
which improves the RHS of \cref{theo: mainRec}.
\end{remark}
\section{Proofs of the Technical Lemmas} \label{sec: prooflemma}

\subsection{Bounding $G_{k,l}, l \geq 1$ (Proof of \cref{lem: concenElke1})}
We expand $ \frac{1}{z^{l-1}} e_1^\top \mathcal{E}^l P$ into $\frac{1}{z^{l-1}} e_1^\top \mathcal{E}^l \left( \sum_{i=1}^r \frac{\sigma_i}{(z-\sigma_i)z} \mathbf{u}_{+i}\mathbf{u}_{+i}^\top+ \sum_{i=1}^{r}\frac{-\sigma_i}{(z+\sigma_i)z} \mathbf{u}_{-i}\mathbf{u}_{-i}^\top\right),$ which can be rearranged as
$$\textstyle  \left[\sum_{i=1}^r \frac{\sigma_i (e_1^\top \mathcal{E}^l \mathbf{u}_{+i}) }{z^l (z-\sigma_i)} \cdot \mathbf{u}_{+i}^\top  \right]+\left[ \sum_{i=1}^r \frac{-\sigma_i (e_1^\top \mathcal{E}^l \mathbf{u}_{-i}) }{z^l (z+\sigma_i)} \cdot \mathbf{u}_{-i}^\top  \right].$$
Since $\{\mathbf{u}_{\pm i} \}_{1 \leq i \leq r}$ are orthonormal system, $\| \frac{1}{z^{l-1}} e_1^\top \mathcal{E}^l P\|_2 $ equals
\begin{equation}
 \textstyle   \frac{1}{|z|^l} \cdot \sqrt{ \sum_{i=1}^r \frac{\sigma_
    i^2 \cdot (e_1^\top \mathcal{E}^l \mathbf{u}_{+i})^2}{|z-\sigma_i|^2}  + \sum_{i=1}^r  \frac{\sigma_i^2 \cdot (e_1^\top \mathcal{E}^l \mathbf{u}_{-i})^2}{|z+\sigma_i|^2}  } \leq \max_{1\leq i \leq r}|e_1^\top \mathcal{E}^l \mathbf{u}_{\pm i}| \cdot \frac{ \sqrt{\sum_{i=1}^r \frac{\sigma_
    i^2}{|z-\sigma_i|^2}  + \sum_{i=1}^r  \frac{\sigma_i^2}{|z+\sigma_i|^2} }  }{|z|^l}
\end{equation}
And hence, 
\begin{equation} \label{ineq: e1EP0}
 \textstyle \frac{1}{2\pi} \cdot \int_{\Gamma} \big\| \frac{1}{z^{l-1}} e_1^\top \mathcal{E}^l P\big\|_2 \,|\, \d z| \leq \frac{\max_{1\leq i \leq r}|e_1^\top \mathcal{E}^l \mathbf{u}_{\pm i}|}{2\pi} \cdot \int_{\Gamma} \frac{ \sqrt{\sum_{i=1}^r \frac{\sigma_
    i^2}{|z-\sigma_i|^2}  + \sum_{i=1}^r  \frac{\sigma_i^2}{|z+\sigma_i|^2} }  }{|z|^l}\,|\, \d z|.
  \end{equation}
The second factor can be bounded as follows. 
\begin{equation}
    \begin{split}
        & \int_{\Gamma} \frac{ \sqrt{\sum_{i=1}^r \frac{\sigma_
    i^2}{|z-\sigma_i|^2}  + \sum_{i=1}^r  \frac{\sigma_i^2}{|z+\sigma_i|^2} }  }{|z|^l}\,|\, \d z|  \leq \sqrt{2r} \int_\Gamma \frac{1 }{|z|^l} \cdot \max_{i \in [r]}\frac{\sigma_i}{|z \pm \sigma_i|} |\, \d z| \\
    & = \sqrt{2r} \int_{\Gamma^{+}} \frac{1 }{|z|^l} \cdot \max_{i \in [r]}\frac{\sigma_i}{|z \pm \sigma_i|} |\, \d z| + \sqrt{2r} \int_{\Gamma^{-}} \frac{1 }{|z|^l} \cdot \max_{i \in [r]}\frac{\sigma_i}{|z \pm \sigma_i|} |\, \d z|.
    \end{split}
\end{equation}
Moreover, 
\begin{equation}  \label{ineq: factor2}
    \sqrt{2r} \int_{\Gamma^{+}} \frac{1 }{|z|^l} \cdot \max_{i \in [r]}\frac{\sigma_i}{|z \pm \sigma_i|} |\, \d z| \leq \sqrt{2r}\frac{\sigma_s}{\delta_s/2} \cdot \int_{\Gamma^{+}} \frac{1}{|z|^l} |\, \d z| \leq \sqrt{2r} \frac{2\sigma_s}{\delta_s} \cdot \frac{1}{(\sigma_s/2)^{l-1}} \cdot 3 \log \left(\frac{10 \sigma_1}{\sigma_s} \right).
\end{equation}
Here, the first follows the fact that $\frac{\sigma_i}{|z-\sigma_i|} \leq \frac{\sigma_i}{|\mathcal{R}e z -\sigma_i|} \leq \frac{\sigma_s}{\delta_s/2},$
while the last inequality is obtained by replacing $\delta_D$ in by \cite{tran2026eigenvalue}{Section 8.2} by $\sigma_s/2$. A similar upper bound for $\sqrt{2r} \int_{\Gamma^{-}} \frac{1 }{|z|^l} \cdot \max_{i \in [r]}\frac{\sigma_i}{|z \pm \sigma_i|} |\, \d z|$. 

Next, we handle the first factor as follows. We split $\mathcal{E}=\mathcal{E}_R+\mathcal{E}_0,$ where $\mathcal{E}_R$ is $(K,\sigma)$-bounded random matrix. By the triangle inequality, we have 
\begin{equation}
    \begin{split}
    |e_1^\top \mathcal{E}^l \mathbf{u}_i| & = |e_1^\top (\mathcal{E}_R+\mathcal{E}_0)^l \textbf{u}_i| \\
    & \leq |e_1^\top \mathcal{E}_R^l \mathbf{u}_i|+\sum_{t=0}^{l-1} \binom{l}{t} \|\mathcal{E}_R\|_{\op}^t \|\mathcal{E}_0\|_{\op}^{l-t}.    \end{split}
\end{equation}
For the first term, by \cite{ltranvufast}[Lemma 4.2], we have
\begin{equation} \label{ineq: factor1}
     \max_{1\leq i \leq r} |e_1^\top \mathcal{E}_R^l \mathbf{u}_{\pm i}| \leq  (2\sigma\sqrt{m+n}) ^l \cdot \left( \bigg[\log (m+n)+ \frac{K \log^3(m+n)}{\sqrt{m+n}} \bigg]\max_{1 \leq i \leq r} \|\mathbf{u}_{\pm i}\|_{\infty} + \frac{\log^{3/2}(m+n)}{\sqrt{m+n}} \right),
\end{equation}
with probability at least $1 - O((m+n)^{-2}) $.

For other terms, using \cite{Vu0, bandeira2016sharp}, with probability at least $1- O((m+n)^{-2})$, we have 
\begin{equation} \label{ineq: factor1extra}
   \binom{l}{t} \|\mathcal{E}_R\|_{\op}^t \|\mathcal{E}_0\|_{\op}^{l-t} \leq \binom{l}{t} \cdot (2\sigma \sqrt{m+n})^t \|\mathcal{E}_0\|_{\op}^{l-t}.  %
\end{equation}
Combing \eqref{ineq: e1EP0}, \eqref{ineq: factor1}, \eqref{ineq: factor1extra}, \eqref{ineq: factor2}, and the line below it, we obtain that $\textstyle \frac{\int_{\Gamma} \big\| \frac{1}{z^{l-1}} e_1^\top \mathcal{E}^l P\big\|_2 \,|\, \d z|}{2\pi C \sqrt{r}} $ is at most
\begin{equation*}
    \begin{split}
       & \big(\frac{4\sigma\sqrt{m+n}}{\sigma_s} \big)^{l-1} \cdot (\sigma \sqrt{m+n} \frac{\sigma_s}{\delta_s}) \bigg[\log^2(m+n)+ \frac{K \log^4(m+n)}{\sqrt{m+n}} \bigg]\max_{1 \leq i \leq r} \|\mathbf{u}_{\pm i}\|_{\infty}  \\
       & + \sum_{t=0}^{l-1}  \binom{l}{t}\big(\frac{4\sigma\sqrt{m+n}}{\sigma_s} \big)^{t} \cdot \left(\frac{2\|\mathcal{E}_0\|_{\op}}{\sigma_s} \right)^{l-t-1} \cdot (\sqrt{m+n} \frac{\sigma_s}{\delta_s}) \cdot \frac{\|\mathcal{E}_0\| \log (m+n)}{\sqrt{m+n}} \\
       & \leq \frac{1}{3^{l-1}}\cdot (\sigma \sqrt{m+n} \frac{\sigma_s}{\delta_s}) \left( \bigg[\log^2(m+n)+ \frac{K \log^4(m+n)}{\sqrt{m+n}} \bigg]\max_{1 \leq i \leq r} \|\mathbf{u}_{\pm i}\|_{\infty} + \frac{\|\mathcal{E}_0\|_{\op} \log (m+n)}{\sigma\sqrt{m+n}} \right),
    \end{split}
\end{equation*}
with probability at least $1 -O((m+n)^{-2})$ for some universal constant $C > 0$.
The last inequality is obtained by the fact that $\sigma_s \geq \delta_s \geq 6(\|\mathcal{E}_R\|_{\op} +\|\mathcal{E}_0\|_{\op}) \geq 3\big[ 4\sigma \sqrt{m+n}+2\|\mathcal{E}_0\|_{\op} \big].$

This proves \cref{lem: concenElke1}.

\subsection{Bounding $G_{k,0}$ (Proof of \cref{lem: concenE0ke1})}
Consider the spectral decomposition of $\mathcal{A} \coloneqq  \sum_{i=1}^n \lambda_j \mathbf{v}_j \mathbf{v}_j^\top,$ where $\lambda_1 \geq \lambda_2 \geq \cdots \geq \lambda_{m+n}$ are its eigenvalues with the corresponding eigenvector $\mathbf{v}_j, 1 \leq j \leq m+n$. It yields another way of presenting $(zI -\mathcal{A})^{-1}= \sum_{j=1}^{m+n} \frac{\mathbf{v}_j \mathbf{v}_j^\top}{z - \mathbf{\lambda}_i}.$ Thus,
    we rewrite $ z e_1^\top P \mathcal{E} (zI- \mathcal{A})^{-1}$ as follows. 
    \begin{equation}
        \begin{split}
  z e_1^\top P \mathcal{E} (zI- \mathcal{A})^{-1} & = \left( \sum_{i=1}^r \frac{\sigma_i \mathbf{u}_{+i1}}{(z-\sigma_i)} \mathbf{u}_{+i}^\top+ \sum_{i=1}^{r}\frac{-\sigma_i \mathbf{u}_{-i1}}{(z+\sigma_i)} \mathbf{u}_{-i}^\top\right) \mathcal{E}  \left(\sum_{j=1}^{m+n} \frac{\mathbf{v}_j \mathbf{v}_j^\top}{z - \mathbf{\lambda}_i} \right)  \\
  & = \sum_{j=1}^{m+n} \left(\sum_{i=1}^r \frac{\sigma_i \mathbf{u}_{+i1}}{(z-\sigma_i)} (\mathbf{u}_{+i}^\top \mathcal{E} \mathbf{v}_j) + \sum_{i=1}^{r}\frac{-\sigma_i \mathbf{u}_{-i1}}{(z+\sigma_i)} (\mathbf{u}_{-i}^\top \mathcal{E} \mathbf{v}_j)  \right) \cdot \frac{\mathbf{v}_j^\top}{z-\mathbf{\lambda}_j}.
        \end{split}
    \end{equation}
Therefore, $\|z e_1^\top P \mathcal{E} (zI- \mathcal{A})^{-1}\|_2$ equals 
$$\sqrt{\sum_{j=1}^{m+n} \left|\sum_{i=1}^r \frac{\sigma_i \mathbf{u}_{+i1}}{(z-\sigma_i)} (\mathbf{u}_{+i}^\top \mathcal{E} \mathbf{v}_j) + \sum_{i=1}^{r}\frac{-\sigma_i \mathbf{u}_{-i1}}{(z+\sigma_i)} (\mathbf{u}_{-i}^\top \mathcal{E} \mathbf{v}_j)  \right|^2 \cdot \frac{1}{|z-\mathbf{\lambda}_j|^2} },$$
which, by Cauchy-Schwartz inequality, is at most 
$$\sqrt{\sum_{j=1}^{m+n} 2r \left( \sum_{i=1}^r \left|\frac{\sigma_i \mathbf{u}_{+i1}}{(z-\sigma_i)} (\mathbf{u}_{+i}^\top \mathcal{E} \mathbf{v}_j) \right|^2 + \sum_{i=1}^{r}\left|\frac{-\sigma_i \mathbf{u}_{-i1}}{(z+\sigma_i)} (\mathbf{u}_{-i}^\top \mathcal{E} \mathbf{v}_j)  \right|^2\right)  \cdot \frac{1}{|z-\mathbf{\lambda}_j|^2} } $$
$$ \leq \max_{\substack{1 \leq i\leq r, 1 \leq j \leq m+n}}|\mathbf{u}_{\pm i}^\top \mathcal{E} \mathbf{v}_j| \cdot \max_{1 \leq i \leq r} \|\mathbf{u}_{\pm i}\|_{\infty} \sqrt{ \sum_{j=1}^{m+n} 2r \big[ \sum_{i=1}^r \frac{\sigma_i^2 }{|z-\sigma_i|^2}  + \sum_{i=1}^{r}\frac{\sigma_i^2}{|z+\sigma_i|^2} \big] \frac{1}{|z-\lambda_j|^2}}.$$
Since $\frac{\sigma_i}{|z-\sigma_i|} \leq \frac{\sigma_i}{|\mathcal{R}e z -\sigma_i|} \leq \frac{\sigma_s}{\delta_s/2}$, $\|z e_1^\top P \mathcal{E} (zI- \mathcal{A})^{-1}\|_2$ is at most 
$$\max_{\substack{1 \leq i\leq r, 1 \leq j \leq m+n}}|\mathbf{u}_{\pm i}^\top \mathcal{E} \mathbf{v}_j| \cdot \max_{1 \leq i \leq r} \|\mathbf{u}_{\pm i}\|_{\infty} \cdot \frac{4r \sigma_s}{\delta_s} \frac{\sqrt{m+n}}{\min_j|z-\lambda_j|}.$$
Thus,
$$\int_{\Gamma} \big\| z e_1^\top P \mathcal{E} (zI- \mathcal{A})^{-1}\big\|_2 \,|\, \d z| \leq \max_{\substack{1 \leq i\leq r\\ 1 \leq j \leq m+n}}|\mathbf{u}_{\pm i}^\top \mathcal{E} \mathbf{v}_j| \cdot \max_{1 \leq i \leq r} \|\mathbf{u}_{\pm i}\|_{\infty} \cdot \frac{4r \sigma_s \sqrt{m+n}}{\delta_s} \int_{\Gamma} \frac{1}{\min_j |z -\lambda_j|} |\, \d z|.$$
By \cite{tran2026eigenvalue}~{Section 8.2}, the RHS is at most 
$$ \max_{\substack{1 \leq i\leq r, 1 \leq j \leq m+n}}|\mathbf{u}_{\pm i}^\top \mathcal{E} \mathbf{v}_j| \cdot \max_{1 \leq i \leq r} \|\mathbf{u}_{\pm i}\|_{\infty} \cdot \frac{4r \sigma_s \sqrt{m+n}}{\delta_s} \cdot 4 \log \left( \frac{10\sigma_1}{\delta_s}\right).$$

\noindent On the other hand, by the Bernstein inequality, with probability at least $1- O((m+n)^{-2})$, 
$$\max_{\substack{1 \leq i\leq r, 1 \leq j \leq m+n}}|\mathbf{u}_{\pm i}^\top \mathcal{E} \mathbf{v}_j| \leq \max_{\substack{1 \leq i\leq r, 1 \leq j \leq m+n}}|\mathbf{u}_{\pm i}^\top \mathcal{E}_R \mathbf{v}_j|+|\mathbf{u}_{\pm i}^\top \mathcal{E}_0 \mathbf{v}_j| = O(\sigma\log n +\|\mathcal{E}_0\|_{\op}).$$
Combining these estimates above, with probability $1-O((m+n)^{-2})$, there is a universal constant $C$, such that 
 $$\int_{\Gamma} \big\| z e_1^\top P \mathcal{E} (zI- \mathcal{A})^{-1}\big\|_2 \,|\, \d z| \leq C\sqrt{r} \log^2(m+n) (\|\mathcal{E}_0\|_{\op}+\sigma) \cdot  ( \sqrt{m+n} \frac{\sigma_s}{\delta_s}) \cdot \max_{1 \leq i \leq r} \|\mathbf{u}_{\pm i}\|_{\infty}.$$
 This proves our lemma.

\subsection{Bounding $\|E_R\|_{\op}$  (Proof of \cref{lem: ER bound})} \label{subsec: proof ER}
Recall the definition that $E_R\coloneqq \tilde{A}^{ub} - \Ex \tilde{A}^{ub} + \tilde{E}^{ub}.$ Our goal is to bound $\|E_R\|_{\op}$. 
By the triangle inequality, 
\begin{equation} \label{Ebound triangle}
    \|E_R\|_{\op} \leq \|\tilde{A}^{ub} - \Ex \tilde{A}^{ub}\|_{\op}+\|\tilde{E}^{ub}\|_{\op} = \|\tilde{A}^{ub} - \Ex \tilde{A}^{ub}\|_{\op}+\|\tilde{E}^{ub}\|_{\op}.
\end{equation}
Moreover, using \cite{bandeira2016sharp}, we obtain
\begin{equation}
    \begin{split}
&\textstyle \|\tilde{A}^{ub} - \Ex \tilde{A}^{ub} \|_{\op} \leq 6 \cdot \left( \max \{M_R, M_C\} + \max_{i,j} \bigg| a_{ij}  \sqrt{\frac{p_{ij}(1-p_{ij})}{p_{i}^2}} \bigg| \cdot \sqrt{\log (m+n)} \right),\,\,\text{where}\,      \\
&\textstyle M_R \coloneqq   \max_{1 \leq i \leq n} \sqrt{ \sum_{j=1}^m a_{ij}^2 \frac{p_{ij}(1-p_{ij})}{p_{i}^2}  }\,\,\text{and}\,\, M_C \coloneqq   \max_{1 \leq j \leq m} \sqrt{ \sum_{i=1}^n a_{ij}^2 \frac{p_{ij}(1-p_{ij})}{p_{i}^2}  }. 
    \end{split}
\end{equation}
Similarly, we also have $\frac{\|\tilde{E}^{ub}\|_{\op} }{6} $ is at most
$$\textstyle  \max_{\substack{1\leq i \leq n \\ 1 \leq j \leq m}} \bigg\{ \sqrt{ \sum_{k=1}^m \Ex \varepsilon_{ik}^2 \frac{p_{ik}(1-p_{ik})}{p_{i}^2}  }, \sqrt{\sum_{l=1}^n \Ex \varepsilon_{lj}^2 \frac{p_{lj}(1-p_{lj})}{p_l^2} }  \bigg\}+\max_{i,j} \bigg| K_E  \sqrt{\frac{p_{ij}(1-p_{ij})}{p_{i}^2}} \bigg| \sqrt{\log (m+n)}.  $$
By the definition of $K_A$ that $\|A\|_{\infty}=\max_{i,j}|a_{ij}|
 \leq K_A$, we have 
 $$\textstyle M_R \leq K_A \cdot \max_{1 \leq i \leq n} \sqrt{ \sum_{j=1}^m \frac{p_{ij}(1-p_{ij})}{p_{i}^2}  } \leq K_A \cdot \max_{1 \leq i \leq n} \sqrt{ \sum_{j=1}^m  \frac{p_{ij}}{p_{i}^2}   } = K_A \cdot \sqrt{\frac{m}{\min_{1\leq i \leq n} p_i}}.$$
 Similarly, 
 $$\textstyle M_C \leq K_A \cdot \max_{1 \leq j \leq m} \sqrt{ \sum_{i=1}^n \frac{p_{ij}(1-p_{ij})}{p_{i}^2}  } \leq K_A \cdot \sqrt{\frac{\max_{1 \leq j \leq m} \sum_{i=1}^n \frac{p_{ij}}{p_i}}{\min_{1 \leq i \leq n}p_i}}.$$
 Therefore, 
 $$\textstyle \max\{M_R, M_C\} \leq K_A \cdot \sqrt{\frac{m+r_pn}{\min p_i}}.$$
 Since 
 $$\textstyle \max_{i,j} \bigg| a_{ij}  \sqrt{\frac{p_{ij}(1-p_{ij})}{p_{i}^2}} \bigg| \leq K_A \cdot \max_{i,j} \sqrt{\frac{p_{ij}/p_i}{p_i}} \leq K_A \cdot \sqrt{\frac{r_p}{\min p_i}},$$
 we obtain
 \begin{equation} \label{Aub bound}
  \textstyle   \|\tilde{A}^{ub} - \Ex \tilde{A}^{ub} \|_{\op} \le 12 K_A \sqrt{\log (m+n)} \cdot \sqrt{\frac{m+r_p n}{\min p_i}}.
 \end{equation}
 By a similar argument, we also obtain
 \begin{equation} \label{Eub bound}
  \textstyle  \|\tilde{E}^{ub}\|_{\op} \leq 12 K_E \sqrt{\log (m+n)} \cdot \sqrt{\frac{m+r_p n}{\min p_i}}. 
 \end{equation}
Combining the inequalities \eqref{Aub bound}, \eqref{Eub bound} with \eqref{Ebound triangle}, we finally obtain
\begin{equation} \label{Ebound}
  \textstyle  \|E_R\|_{\op} \leq 12 (K_A+K_E) \bigg( \sqrt{\log (m+n)} \cdot \sqrt{\frac{m+r_p n}{\min p_i}} \bigg).
\end{equation}
We complete the proof of \cref{lem: ER bound}.

\end{document}

%% file: headers/head.tex
\input{./headers/bbhead}

\input{./headers/calhead}
\input{./headers/lrhead}

\input{./headers/vecmathead}
\input{./headers/thmhead}

\input{./headers/comphead}

\input{./headers/latthead}


\input{./headers/marginnotes}

\input{./headers/probhead}
\input{./headers/math}

\mathchardef\mdash="2D %

\renewcommand{\epsilon}{\varepsilon}

\def\compactify{\itemsep=0pt \topsep=0pt \partopsep=0pt \parsep=0pt}
\let\latexusecounter=\usecounter

{\begin{itemize}%
\setlength{\itemsep}{0pt}%
\setlength{\topsep}{0pt}%
\setlength{\partopsep}{0 in}%
\setlength{\parskip}{0 pt}}%
{\end{itemize}}

\definecolor{niceRed}{RGB}{190,38,38}
\definecolor{niceYellow}{HTML}{f5b400}
\definecolor{blueGrotto}{HTML}{059DC0}
\definecolor{royalBlue}{HTML}{057DCD}
\definecolor{navyBlue}{HTML}{0B579C}
\definecolor{yaleBlue}{HTML}{00356b}
\definecolor{limeGreen}{HTML}{81B622}
\definecolor{nicePurple}{HTML}{9c27b0}
\definecolor{lightRoyalBlue}{HTML}{def2ff}  
\definecolor{gold}{HTML}{ffa300}

\renewcommand{\theequation}{\thesection.\arabic{equation}}

%% file: headers/bbhead.tex
\newcommand{\N}{\ensuremath{\mathbb{N}}}

\newcommand{\R}{\ensuremath{\mathbb{R}}}

%% file: headers/lrhead.tex
\NewDocumentCommand\xDeclarePairedDelimiter{mmm}
 {%
  \NewDocumentCommand#1{som}{%
   \IfNoValueTF{##2}
    {\IfBooleanTF{##1}{#2##3#3}{\mleft#2##3\mright#3}}
    {\mathopen{##2#2}##3\mathclose{##2#3}}%
  }%
 }

\xDeclarePairedDelimiter\inner{\langle}{\rangle}
\xDeclarePairedDelimiter\ang{\langle}{\rangle}
\xDeclarePairedDelimiter\abs{\lvert}{\rvert}
\xDeclarePairedDelimiter\set{\{}{\}}
\xDeclarePairedDelimiter\p{(}{)}

\let\b=\relax
\xDeclarePairedDelimiter\b{[}{]}
\xDeclarePairedDelimiter\round{\lfloor}{\rceil}
\xDeclarePairedDelimiter\floor{\lfloor}{\rfloor}
\xDeclarePairedDelimiter\ceil{\lceil}{\rceil}
\xDeclarePairedDelimiter\norm{\lVert}{\rVert}

%% file: headers/thmhead.tex
\theoremstyle{plain}            %
\newtheorem{theorem}{Theorem}[section]
\newtheorem{lemma}[theorem]{Lemma}
\newtheorem{corollary}[theorem]{Corollary}

\newtheorem{infassumption}[theorem]{Informal Assumption}

\theoremstyle{definition}       %
\newtheorem{definition}[theorem]{Definition}

\newtheorem{assumption}[theorem]{Assumption}

\theoremstyle{remark}           %
\newtheorem{remark}[theorem]{Remark}

\numberwithin{equation}{section}

\AfterEndEnvironment{definition}{\noindent\ignorespaces}
\AfterEndEnvironment{infdefinition}{\noindent\ignorespaces}
\AfterEndEnvironment{assumption}{\noindent\ignorespaces}
\AfterEndEnvironment{problem}{\noindent\ignorespaces}
\AfterEndEnvironment{openproblem}{\noindent\ignorespaces}
\AfterEndEnvironment{lemma}{\noindent\ignorespaces}
\AfterEndEnvironment{theorem}{\noindent\ignorespaces}
\AfterEndEnvironment{proposition}{\noindent\ignorespaces}
\AfterEndEnvironment{fact}{\noindent\ignorespaces}
\AfterEndEnvironment{question}{\noindent\ignorespaces}
\AfterEndEnvironment{corollary}{\noindent\ignorespaces}
\AfterEndEnvironment{model}{\noindent\ignorespaces}
\AfterEndEnvironment{remark}{\noindent\ignorespaces}
\AfterEndEnvironment{proof}{\noindent\ignorespaces}
\AfterEndEnvironment{fact}{\noindent\ignorespaces}
\AfterEndEnvironment{minftheorem}{\noindent\ignorespaces}
\AfterEndEnvironment{inftheorem}{\noindent\ignorespaces}
\AfterEndEnvironment{maintheorem}{\noindent\ignorespaces}
\AfterEndEnvironment{restatable}{\noindent\ignorespaces}
\AfterEndEnvironment{infassumption}
{\noindent\ignorespaces}
\AfterEndEnvironment{infcorollary}{\noindent\ignorespaces}

\crefname{section}{Section}{Sections}
\crefname{theorem}{Theorem}{Theorems}
\crefname{lemma}{Lemma}{Lemmas}
\crefname{problem}{Problem}{Problems}
\crefname{program}{Program}{Progams}
\crefname{definition}{Definition}{Definitions}
\crefname{conjecture}{Conjecture}{Conjectures}
\crefname{corollary}{Corollary}{Corollaries}
\crefname{construction}{Construction}{Constructions}
\crefname{conjecture}{Conjecture}{Conjectures}
\crefname{claim}{Claim}{Claims}
\crefname{observation}{Observation}{Observations}
\crefname{proposition}{Proposition}{Propositions}
\crefname{fact}{Fact}{Facts}
\crefname{question}{Question}{Questions}
\crefname{problem}{Problem}{Problems}
\crefname{remark}{Remark}{Remarks}
\crefname{example}{Example}{Examples}
\crefname{equation}{Equation}{Equations}
\crefname{appendix}{Section}{Sections}
\crefname{algorithm}{Algorithm}{Algorithms}
\crefname{model}{Model}{Models}
\crefname{figure}{Figure}{Figures}
\crefname{infassumption}{Informal Assumption}{Informal Assumptions}
\crefname{inftheorem}{Informal Theorem}{Informal Theorems}
\crefname{infdefinition}{Informal Definition}{Informal Definitions}
\crefname{minftheorem}{Main Informal Theorem}{Main Informal Theorems}
\crefname{maintheorem}{Main Theorem}{Main Theorems}
\crefname{assumption}{Assumption}{Assumptions}
\crefname{step}{Step}{Steps}
\crefname{result}{Result}{Results}
\crefname{event}{Event}{Events}
\crefname{none}{}{}

\usepackage{enumitem}

\newlist{asmpenum}{enumerate}{1} %
\setlist[asmpenum]{label={\arabic*.},ref=\theassumption.{\arabic*}}
\crefname{asmpenumi}{Assumption}{Assumptions}

\contourlength{0.1pt}
\contournumber{10}

%% file: headers/comphead.tex
\DeclareMathOperator*{\wt}{wt}

\def\poly{\mathrm{poly}}

%% file: headers/latthead.tex
\DeclareMathOperator{\rank}{rank}

%% file: headers/marginnotes.tex
\newif\ifnotes\notestrue

\ifnotes
\usepackage{color}
\definecolor{mygrey}{gray}{0.50}
\newcommand{\notename}[2]{{\textcolor{red}{\footnotesize{\bf (#1:} {#2}{\bf
) }}}}

\else

\newcommand{\notename}[2]{{}}

\fi